\documentclass[3p,final,12pt]{elsarticle}
\usepackage{diagbox}
\usepackage{booktabs}
\usepackage{siunitx}
\usepackage{tabulary,xcolor}
\usepackage{amsfonts,amsmath,amssymb}
\usepackage[T1]{fontenc}

\biboptions{sort&compress}

\usepackage[linesnumbered,ruled,vlined]{algorithm2e}
\SetKwInput{KwInput}{Input}                
\SetKwInput{KwOutput}{Output}              

\usepackage{ifluatex}
\ifluatex
\usepackage{fontspec}
\defaultfontfeatures{Ligatures=TeX}
\usepackage[]{unicode-math}
\unimathsetup{math-style=TeX}
\else 
\usepackage[utf8]{inputenc}
\fi 
\ifluatex\else\usepackage{stmaryrd}\fi

\usepackage{url,multirow,morefloats,floatflt,cancel,tfrupee}
\makeatletter

\AtBeginDocument{\@ifpackageloaded{textcomp}{}{\usepackage{textcomp}}}
\makeatother
\usepackage{colortbl}
\usepackage{xcolor}
\usepackage{pifont}
\usepackage[nointegrals]{wasysym}
\makeatletter

\def\mcWidth#1{\csname TY@F#1\endcsname+\tabcolsep}

\def\cAlignHack{\rightskip\@flushglue\leftskip\@flushglue\parindent\z@\parfillskip\z@skip}
\def\rAlignHack{\rightskip\z@skip\leftskip\@flushglue \parindent\z@\parfillskip\z@skip}

\@ifundefined{etal}{}{}

\usepackage{ifxetex}
\ifxetex\else\if@twocolumn\@ifpackageloaded{stfloats}{}{\usepackage{dblfloatfix}}\fi\fi

\AtBeginDocument{
\expandafter\ifx\csname eqalign\endcsname\relax
\def\eqalign#1{\null\vcenter{\def\\{\cr}\openup\jot\m@th
  \ialign{\strut$\displaystyle{##}$\hfil&$\displaystyle{{}##}$\hfil
      \crcr#1\crcr}}\,}
\fi
}

\AtBeginDocument{%
  \@ifpackageloaded{endfloat}%
   {\renewcommand\efloat@iwrite[1]{\immediate\expandafter\protected@write\csname efloat@post#1\endcsname{}}}{\newif\ifefloat@tables}%
}%

\def\BreakURLText#1{\@tfor\brk@tempa:=#1\do{\brk@tempa\hskip0pt}}
\let\lt=<
\let\gt=>
\def\processVert{\ifmmode|\else\textbar\fi}

\@ifundefined{subparagraph}{
\def\subparagraph{\@startsection{paragraph}{5}{2\parindent}{0ex plus 0.1ex minus 0.1ex}%
{0ex}{\normalfont\small\itshape}}%
}{}

\newcommand\role[1]{\unskip}
\newcommand\aucollab[1]{\unskip}
  
\@ifundefined{tsGraphicsScaleX}{\gdef\tsGraphicsScaleX{1}}{}
\@ifundefined{tsGraphicsScaleY}{\gdef\tsGraphicsScaleY{.9}}{}
\def\checkGraphicsWidth{\ifdim\Gin@nat@width>\linewidth
	\tsGraphicsScaleX\linewidth\else\Gin@nat@width\fi}

\def\checkGraphicsHeight{\ifdim\Gin@nat@height>.9\textheight
	\tsGraphicsScaleY\textheight\else\Gin@nat@height\fi}

\def\fixFloatSize#1{}
\let\ts@includegraphics\includegraphics

\def\inlinegraphic[#1]#2{{\edef\@tempa{#1}\edef\baseline@shift{\ifx\@tempa\@empty0\else#1\fi}\edef\tempZ{\the\numexpr(\numexpr(\baseline@shift*\f@size/100))}\protect\raisebox{\tempZ pt}{\ts@includegraphics{#2}}}}

\AtBeginDocument{\def\includegraphics{\@ifnextchar[{\ts@includegraphics}{\ts@includegraphics[width=\checkGraphicsWidth,height=\checkGraphicsHeight,keepaspectratio]}}}

\DeclareMathAlphabet{\mathpzc}{OT1}{pzc}{m}{it}

\def\URL#1#2{\@ifundefined{href}{#2}{\href{#1}{#2}}}

\def\UrlOrds{\do\*\do\-\do\~\do\'\do\"\do\-}%
\g@addto@macro{\UrlBreaks}{\UrlOrds}

\usepackage[
singlelinecheck=false 
]{caption}
\edef\fntEncoding{\f@encoding}

\makeatother

\newif\ifmultipleabstract\multipleabstractfalse%
\usepackage{float}

\makeatletter
\def\ps@pprintTitle{%
 \let\@oddhead\@empty
 \let\@evenhead\@empty
 \def\@oddfoot{}%
 \let\@evenfoot\@oddfoot}
\makeatother

\begin{document}

\begin{frontmatter}
	
\title{Data-driven Identification of 2D Partial Differential Equations using extracted physical features
}
    
\author[1]{Kazem Meidani}
\author[1,2,3]{Amir Barati Farimani\corref{main}}\cortext[main]{Corresponding author}
\ead{barati@cmu.edu}
    
\address[1]{Department of Mechanical Engineering\unskip, 
    Carnegie Mellon University, Pittsburgh, PA, USA}
\address[2]{Department of Chemical Engineering\unskip, 
    Carnegie Mellon University, Pittsburgh, PA, USA}
\address[3]{Machine Learning Department\unskip, 
Carnegie Mellon University, Pittsburgh, PA, USA}
\begin{abstract}
Many scientific phenomena are modeled by Partial Differential Equations (PDEs). The development of data gathering tools along with the advances in machine learning (ML) techniques have raised opportunities for data-driven identification of governing equations from experimentally observed data. We propose an ML method to discover the terms involved in the equation from two-dimensional spatiotemporal data. Robust and useful physical features are extracted from data samples to represent the specific behaviors imposed by each mathematical term in the equation. Compared to the previous models, this idea provides us with the ability to discover 2D equations with time derivatives of different orders, and also to identify new underlying physics on which the model has not been trained. Moreover, the model can work with small sets of low-resolution data while avoiding numerical differentiations. The results indicate robustness of the features extracted based on prior knowledge in comparison to automatically detected features by a Three-dimensional Convolutional Neural Network (3D CNN) given the same amounts of data. Although particular PDEs are studied in this work, the idea of the proposed approach could be extended for reliable identification of various PDEs.
\end{abstract}

\end{frontmatter}

\section{Introduction}
  
Partial Differential Equations (PDEs) have been pivotal in providing comprehensive and accurate models for many scientific phenomena, specifically in various disciplines of physics and engineering. Navier-Stokes equations in fluid dynamics, Fourier's heat conduction equation, Maxwell's equations in electromagnetism, and Schr\"{o}dinger's equation in quantum mechanics are some examples of PDEs that govern physical world phenomena. 
  
Originally, these models were discovered with the awareness of theory, mathematical modeling, and supports from observations. In fact, seemingly distinct physical processes are modeled by equations that are similar in being composed of some usually meaningful mathematical components. Physical behaviors imposed by different parts of the model on the observed data could help experts to identify the model with prior knowledge of theory and data. 
  
  Recent developments in data collection, storage, and processing tools enable us to access huge amounts of data from experiments and simulations. This availability paves the way for using data-driven methods to extract knowledge from raw data. In the past decades, Machine Learning (ML) techniques have revolutionized many fields like computer vision, and they can be helpful in solving and understanding physical problems when being applied to scientific data. 
  
  Data-driven methods have been applied to experimentally observed data to model, predict, or control systems. Various Deep Learning (DL) algorithms have been used for learning time-dependent systems from training data to solve PDEs by approximating the solution given the system conditions \cite{solvePDE-Han, DGM, deepxde} or to model the phenomena \cite{wiewel2018latentspace,farimani2017deep,LSTMchaos,kim_lee_2020}.  Prior knowledge of the system, like the equation or basic laws of physics, could also be incorporated to enhance the learning process and make the model more stable and robust \cite{PINN-inverse,CPINN,DL_PhysicalProcesses,PI-deep-generative,weaklysupervised-Barati,hamiltonian}.

  Besides modeling for solution and prediction, there have been great efforts to discover the underlying physics of systems from data \cite{Crutchfield1987}. A capable mathematical model should keep balance of accuracy and complexity. This idea was used in an evolutionary algorithm to determine the governing equation of dynamical systems \cite{Bongard-Lipson,Schmidt-Lipson}. Brunton et al. \cite{SINDY} made use of sparse regression framework to discover dynamical systems. In this approach, first, numerical methods are applied to the time dependent data to build a large dictionary of terms that are highly possible to be present in the equation. Assuming a general form of equation, sparsity promoting loss functions and thresholdings are integrated into regression to moderate the complexity. Sparse optimization has been then used for data-driven discovery of PDEs \cite{PDE-Find,Schaeffer,Parametric-PDE}. These methods are highly interpretable since the explicit form of the equation could be identified, and they can be applied to a vast variety of PDEs. However, complex or high degree equations cannot be discovered due to the discrete and limited choice of terms in the library.  Building a library of functional forms and spatial derivatives requires numerical differentiation which is unstable, and thus errors will accumulate barring the data are clean and from finite difference grids. This makes the method brittle in the presence of noisy experimental data or simulated data with schemes like Finite Element Method (FEM). Incorporating data assimilation methods \cite{Chang-2019} and Gaussian process regression, given the known PDE form, can improve the model in the presence of noise \cite{raissi-Gaussian,hidden-physics}.

  Artificial Neural Networks (ANNs) have been known as universal smooth function approximators \cite{universal-f-apprx} provided proper architecture and training method. Recent studies have leveraged deep learning abilities to identify equations in complex or noisy systems \cite{discovery-complex, dl-pde}. Inspired from their previous work on physics-informed deep learning \cite{PINN-inverse}, Raissi et al. employed multi-step time-stepping schemes as loss function to learn nonlinear dynamical systems \cite{raissi-multistep}. This framework can learn more complex systems than parsimonious methods but at the cost of losing interpretability. Deep learning of hidden physics models is extended to discover nonlinear PDEs from scattered spatiotemporal data \cite{DeepHiddenPhysics}. A wide variety of nonlinear PDEs can be learned as black box models to be further used for solution or prediction of the system. Automatic differentiation in neural networks facilitates calculation of derivatives and nonlinear terms via avoiding numerical differentiation. Long et al. made use of convolutional layers for numerical approximation of differential operators with properly constrained filters which grant expressive power to the model. Therefore, a symbolic neural network can be used to discover the equation form \cite{PDE-Net,PDE-Net-2.0}.

    All the discussed discovery methods aim to mathematically uncover the PDE models without prior knowledge of the systems. The approach shared among these studies usually consists of three steps. First, the functional forms of mathematical terms and derivatives are computed by various differentiation methods. Second, the model is assumed to be of the general form as in Equation 1 with differences in the parameters. Finally, the proposed framework is used to find the function $\mathcal{F}$ whether explicitly or implicitly. The presumed form hinders these methods to simply identify equations that are not time-dependent, e.g. Laplace equation, or have higher orders of time derivatives, e.g. wave equation. The frameworks including sparse regression and neural networks have to be tailored or trained on data from each system. Besides, two-dimensional PDEs are not covered in some of these approaches. 
    \begin{equation}
    u_{t} = \mathcal{F} (u,u_{x},u_{xx},...)
    \end{equation}

  A wide variety of scientific phenomena with known models are well-studied and the physical behavior of solutions could be described by the mathematical form of solution and the PDE model. The key idea of this work is to exploit prior knowledge of known mathematical models to identify PDEs from data. As opposed to relying solely on mathematical frameworks, which result in relatively brittle models, integrating known physics leads to more stable and robust models. Previous models had to deal with differentiation to find the relations, however, features could be extracted from averaged behavior of data over time or at specific time steps. By feature extraction, the data is usually reduced to a scalar or vector which can describe a specific characteristic of data. In this paper, we integrate the knowledge of PDEs to find useful features that explain the solution and thus can be used by a machine learning model to identify the corresponding equation. 
  
  Without the necessity for equation form assumption, we can study equations with various orders of time derivative. Examples are Laplace equation, as a time-independent PDE, and wave equation as a PDE with second derivative with respect to time. PDEs studied in this work are two-dimensional in space that cover a lot of phenomena in nature or engineering, while most of the former methods are designed for one-dimensional equations. Moreover, to identify the equation of a new observation in preceding studies, the data should undergo numerical preprocessing or a time-consuming process of network training. In this paper, however, new observation could be easily fed to the trained model to find the governing structure.

\section{Method}
  The main idea of this work is to leverage prior physical knowledge in combination with data-driven techniques. We show that a model can be trained with observations of known PDEs to learn and identify new observed samples. Useful information embedded in the raw data cannot be directly exploited to detect the presence and extent of specific physical behaviors such as diffusivity in a heat conduction solution. Prior insight about physics and theory of equations is utilized to extract beneficial information from raw data. Finally, the data-driven model makes use of these purified features to identify the building blocks of the equation. In this section, first, we review the equations examined in this study and their expounded.
  
 \subsection{Physical characteristics in PDEs}
  In this paper, Partial Differential Equations with the general form of Equation 2a are considered. If the coefficient of each term is zero, the corresponding term is removed from the equation. 
  
 \begin{subequations}
    \begin{equation}
    eu_{tt} + du_{t} - c \nabla^{2}u + B\nabla u = 0
    \end{equation}  
    \begin{equation}
    u_{t} - c \nabla^{2}u + B\nabla u = 0
    \end{equation}  
\end{subequations}

  Each of the present terms in the equation has a physical meaning which explains their physical effect in the solution. In the case of $e=0$, the equation would be reduced to equation 2b which is the general form of convection-diffusion equations. Therefore, $ c \nabla^{2}u $ indicates diffusion term and $B\nabla u$ is responsible for convection in direction $\vec{B}$.
  On the other hand, if $e\neq 0$, the term $du_{t}$ turns to the damping term which physically reduces the energy of the system over time. The coefficients corresponding to the terms along with their physical meanings are summarized in table 1. 
  
\begin{table}[htbp]
\begin{center}
\caption{Physical meaning of terms in equations}
$eu_{tt} + du_{t} - c \nabla^{2}u + B\nabla u = 0$
{\begin{tabular}{llll}    
\hline
$eu_{tt} = 0$             & Convection-Diffusion        & $eu_{tt}\neq 0$           & Wave  \\
\hline
 $e$& $-$  &$e$ & Mass coefficient    \\
 $d$& Mass coefficient      & $d$  &   Damping coefficient \\
 $c$& Diffusion coefficient &  $c$& Propagation speed  \\ 
 $B$& Convection coefficient &  $B$& Convection coefficient  \\
\hline
\end{tabular}}
\end{center}
\end{table}
  
  The presence or absence of each of these terms in the PDE determines whether the solution to the equation would possess a specific behavior or not. Each PDE, based on the aforementioned general form, models a two-dimensional spatiotemporal phenomena $u(x,y,t)$. Since time direction is fundamentally different from spatial directions, we divide the terms into two categories of time derivatives and space derivatives.

\subsection{Feature Extraction}
  An adequate extracted feature should possess some important properties. First, it should be robust, which leads to the generalizability of the model. In Convolutional Neural Networks (CNNs), features are usually detected automatically from the data. Adversarial examples showed that non-robust features could damage the generalizability of the model when confronting slightly noisy data \cite{goodfellow2014-adversarial,adversarial-NIPS2019}. However, robust useful features are conceptually interpretable and meaningful which would result in the transferability \cite{transferability-adv} of the constructed model. Second, a good feature should be useful in carrying out the desired task of the model. For example in classification tasks, features by which the model can distinguish between different classes are of higher importance in comparison to the features that are common among data samples.  
 
\subsubsection{Time derivatives}
   
   Variations of the system over time show its time dependent behavior. Time derivatives could be approximated by differentiation of the system in two close time steps. Therefore, we can assume that the subtraction of every consecutive time steps can represent the time dependency of the system. Since the focus of this feature is time dependent behavior, we can average the solution $u$ in space for each time step (Equation 3). As a result, this process provides us with a one-dimensional array of size $T-1$, with $T$ being the number of time steps. Figure 1a shows the time dependent behavior of the extracted signal for some data samples with different governing PDEs. In Laplace equation, the solution is not time dependent, hence, there is no change in the system through time. In heat conduction equation, all the singularities in the system are lost immediately, and the information will be lost gradually over time. Therefore, diffusion equations will smoothly converge to steady state after a while. On the contrary, in wave equation, with the second derivative of time, the information is transported and the solution does not decay to steady state unless it is being damped. However, even in damped wave equation, the signal of the change in consecutive time steps is not smooth anymore as it does not correspond to the energy of the system. 
   
    \begin{equation}
    \Delta u(t) = \frac{\sum_{i}^{\mathcal{N}_{x}} \sum_{j}^{\mathcal{N}_{y}} (u^{t+1}_{i,j} - u^{t}_{i,j})}{\mathcal{N}_{x}\mathcal{N}_{y}} = Mean_{x,y \in \mathcal{D}} (u^{t+1}_{i,j} - u^{t}_{i,j})  ,   t=1,2,...,T-1
    \end{equation}  
    
\begin{figure*}[!tpb]    
\begin{center}
  \includegraphics[width=1\linewidth]{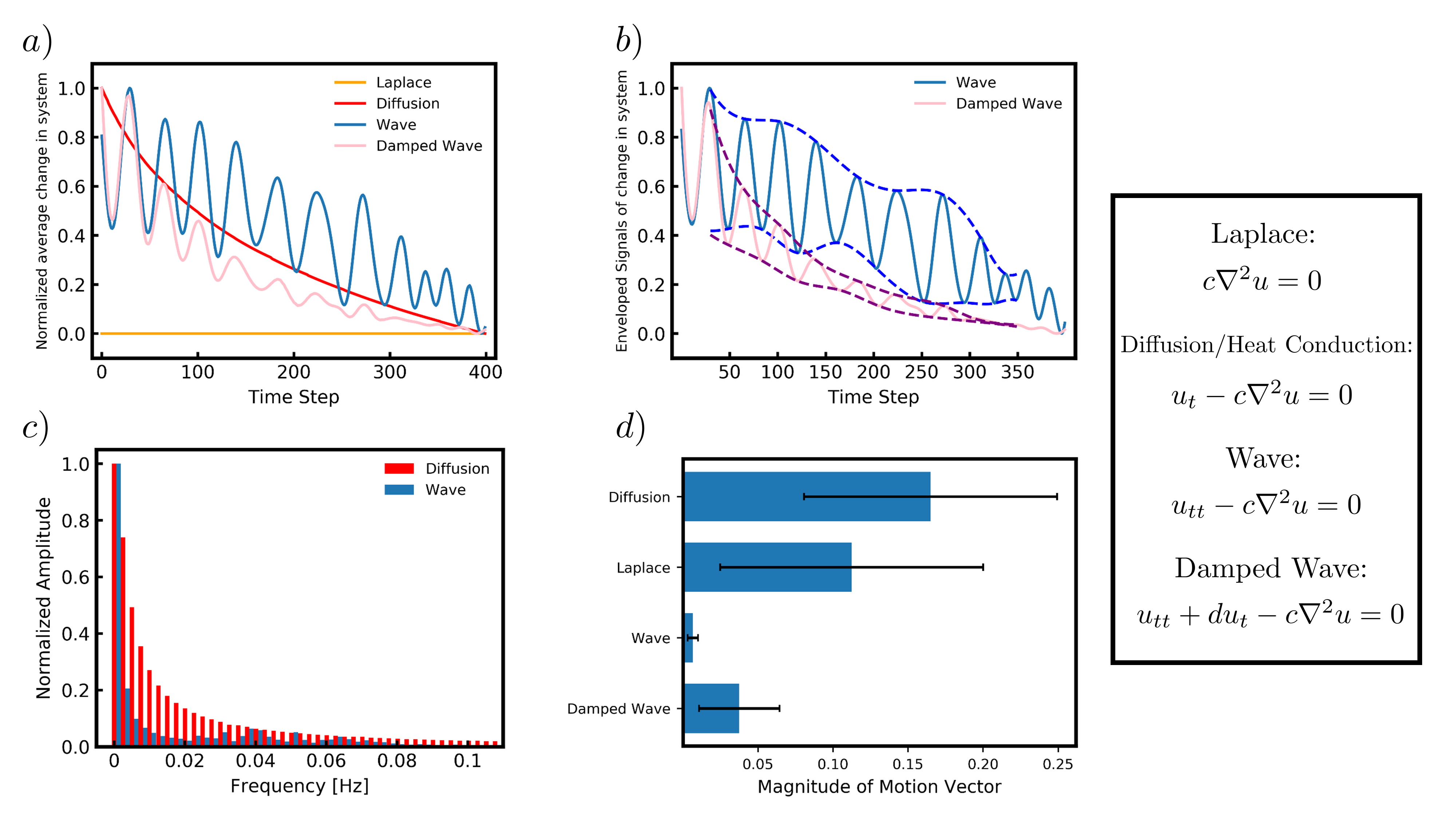}
\end{center}
  \caption{  \textbf{a)} Time signal of the change in consecutive time steps vs. time for one sample of some PDEs (Equations are given in the box). \textbf{ b)} Time signal with envelopes to illustrate the change in the amplitude of time signals. Results are shown for one representative sample of equations. \textbf{c)} Amplitude vs. Frequency for Fast Fourier Transform of a sample of Diffusion and Wave equations. \textbf{d)} Mean magnitude of motion vector averaged over all data samples for different PDEs. Though errors are relatively high, the differences among equations make the feature useful.}
\label{fig:long}
\end{figure*}

   Some informative features can be extracted based on the time dependent and averaged change of the solution, which is viewed as a signal. Since the behavior of the signal, and not the exact values, is the useful feature that helps to distinguish PDEs, we first normalize them to the range of $[0,1]$. As the next step, since the data may contain a little intrinsic noise, Savitzky-Golay smoothing filter \cite{savitzky} is applied to smoothen the signals (\textcolor{blue}{SI section 1.1}).
  
  \begin{itemize}
   \item Statistics of the signal
   
   Statistical moments are basic features that can describe a data or signal. To this end, mean, standard deviation, and skewness of sections of these time signals are extracted for all of the samples (\textcolor{blue}{SI section 1.2}). 
   
   \item Oscillations and Amplitude of the signal 
   
   The oscillating behavior and the amplitude, i.e. peak to valley distance, of the time signals over time can represent the changes in the energy of the system. Hence, the upper and lower envelopes of each signal, are used to examine these properties. Particularly, we may detect damping term as it causes the decay of amplitude (Fig. 1b, \textcolor{blue}{SI section 1.3}). 

   \item Frequency analysis of the signal 
   
   Fourier analysis of the signals using Fast Fourier Transform (FFT) can generate applicable features as it reveals information about the frequency domain of the signal and its oscillations. Normalized FFT signals from various systems show different frequency characteristics (Fig. 1c, \textcolor{blue}{SI section 1.4}).
   
   \item Motion magnitude 

    Unlike former features that were averaged over the spatial domain, here we examine more detailed changes in the system. The method is based on motion detection at each time step. Consider the solution $u^{t}_{i,j}$ on a grid point is moved by a vector $\vec{v}$ in time $\delta t$. Assuming that system change in two consecutive time steps is relatively small, we can only inspect a small filter of size $k\times k$ around the specific block at time $t+1$ to find the vector $\vec{v}$ toward the position with minimum error \cite{motion-detection}. The filter strides and covers the whole space and is averaged over all time steps (Fig. 2). The vector is not allowed to be zero, i.e. to select the same position. So, for solutions with steady or uniform motions, vectors in the same direction would be accumulated over time and the magnitude of averaged motion vector would be higher (Fig. 1d, \textcolor{blue}{SI section 1.5}). 
    

\begin{figure*}[!htpb]    
\begin{center}
  \includegraphics[width=0.8\linewidth]{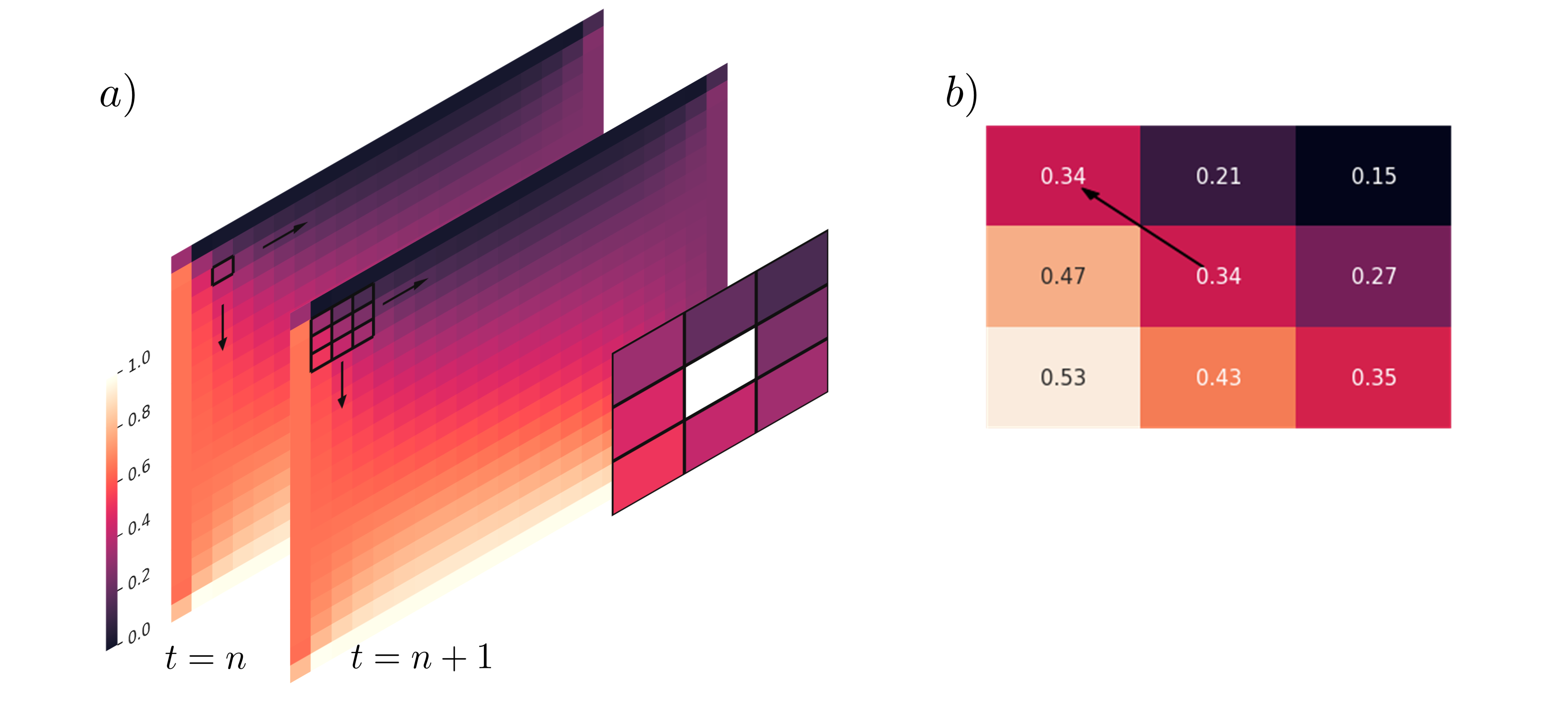}
\end{center}
  \caption{  \textbf{a)} Data in two time steps with the moving $3\times 3$ filter \textbf{b)} Vector is toward the grid point at time $t+1$ with the nearest value to the original grid at time $t$ }
\label{fig:long}
\end{figure*}

   \end{itemize}
   \subsubsection{Space derivatives} 
   
   While the most important features for detection of time derivatives are related to the spatially averaged properties, the distribution of solution in different space directions is the most informative feature about space derivatives. Concerning PDEs covered in this study, the directions of motion prompted by diffusion and convection have some basic differences. Diffusion and wave propagation are triggered by internal gradients of concentration, temperature, etc. which are strongly dependent on the initial and boundary conditions. However, convection is physically a forced motion that its direction is determined by the coefficients. Therefore, we can discover the overall motion direction in the data samples as a feature indicator of spatial derivatives.
   
   We can analyze the normalized solution at time step $t$ on a square domain $\mathcal{D}_{x,y}$ with the assumption of approximate Dirichlet boundary conditions applied on this domain. Figure 3a shows the heat map of the data snapshot for two equations that have differences in the convection term. Next, the distribution of the solution in $x,y$ directions is extracted (equation 4). Given solution boundary and initial conditions, a new coordinate system of the overall motion direction and its perpendicular direction $[\vec{v_2},\vec{v_1}]$ could be calculated from solution gradient $\Delta \mathcal{C}$ in each direction (Equation 5). Finally, we project the spatial distribution of solution into the direction $\vec{v_1}$ (Fig. 3b). Theoretically, if the motion is solely created by $c\nabla^2 u$, we should observe a symmetrical behavior in Figure 3c, and if convection exists, this state will not hold true (\textcolor{blue}{SI section 1.6}). 
   
    \begin{equation}
    \bar{u}_{i,x-dir} = \frac{\sum_{j}^{\mathcal{N}_{y}} u^{t}_{i,j}}{\mathcal{N}_{y}}\quad , \quad
    \bar{u}_{j,y-dir} = \frac{\sum_{i}^{\mathcal{N}_{x}} u^{t}_{i,j}}{\mathcal{N}_{x}}
    \end{equation}  

    \begin{subequations}
    \begin{equation}
    \begin{aligned}
    d_1 = \frac{\Delta{\mathcal{C}_x}}{\| \Delta{\mathcal{C}} \|_2} \quad ,\quad
    d_2 = \frac{\Delta{\mathcal{C}_y}} { \| \Delta{\mathcal{C}} \|_2}
    \end{aligned}
    \end{equation}
    \begin{equation}
    \begin{aligned}
    v_1 = d_1 \bar{u}_y - d_2 \bar{u}_x
      \\
    v_2 = d_1 \bar{u}_x + d_2 \bar{u}_y
    \end{aligned}
    \end{equation}
    \end{subequations}

    \begin{figure*}[!htpb]    
    \begin{center}
      \includegraphics[width=1\linewidth]{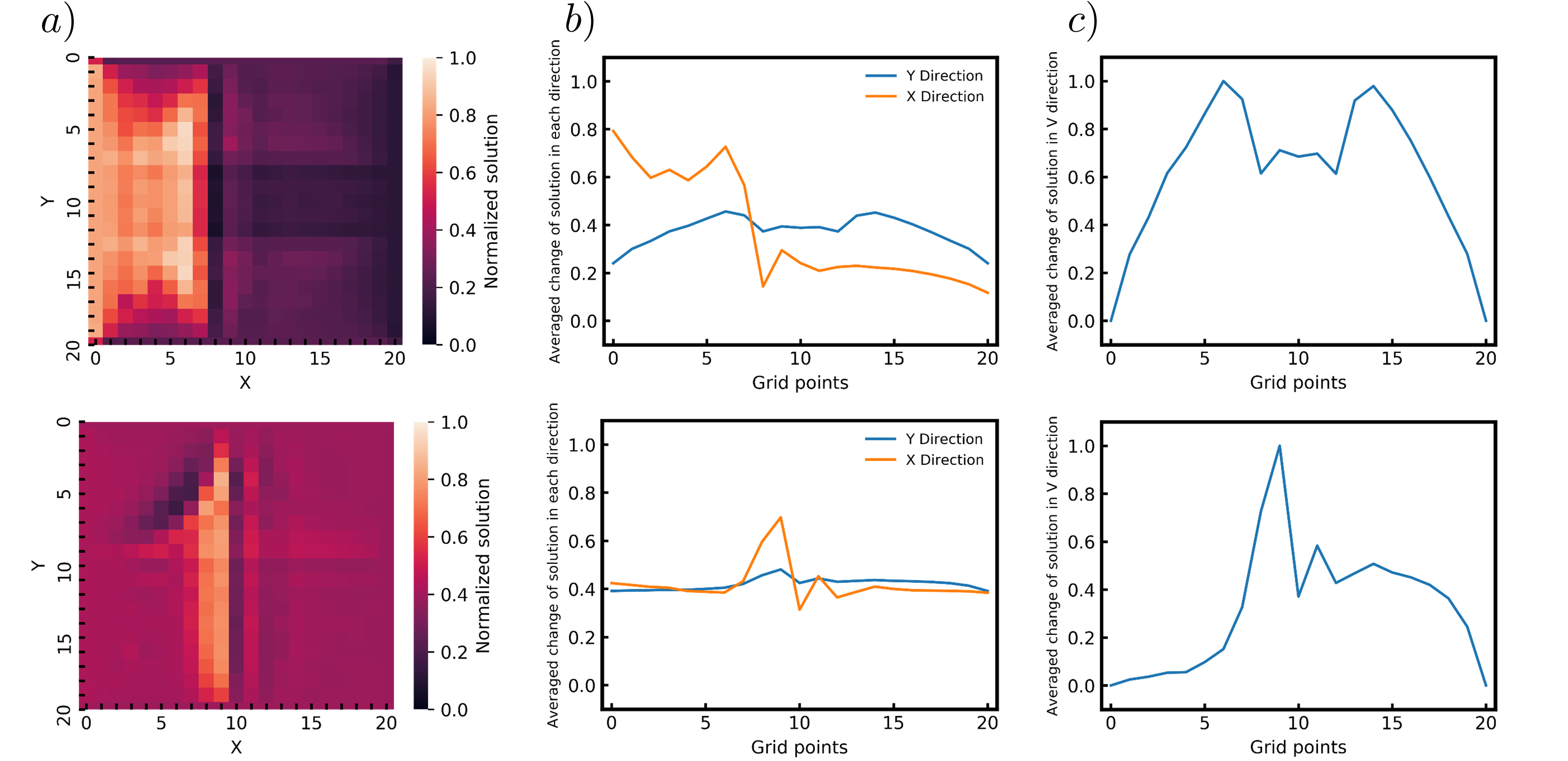}
    \end{center}
      \caption{ Space derivative feature extraction process for (top) $u_{tt} - c\nabla ^{2} u = 0$ (bottom) $u_{tt} - c\nabla ^{2} u + \vec{B}\vec{\nabla} u = 0$ \textbf{a)} Snapshot of normalized data at time step $t$ \textbf{b)} Variation of data in $X,Y$  directions. \textbf{c)} Variation of data in $v_{1}$ direction. Asymmetry of the signal can reveal convection term.}
    \label{fig:long}
    \end{figure*}

   \subsection{Data-driven model}
   
We make use of XGBoost algorithm \cite{xgboost} as a powerful classifier to fit a model for each mathematical term based on the extracted features from observations. With different supervision labels, we can categorize the data not only to different PDEs but also to equations with or without a specific term involved. For example, we can fit a model to classify equations in which $eu_{tt}=0$ vs. equations with  $eu_{tt}\neq0$. We can identify the equation by combining these models in a pipeline (Fig. 4). We use scikit-learn python module for the implementation of the model \cite{scikit-learn}.
    
\begin{figure*}[!htpb]    
\begin{center}
  \includegraphics[width=0.9\linewidth]{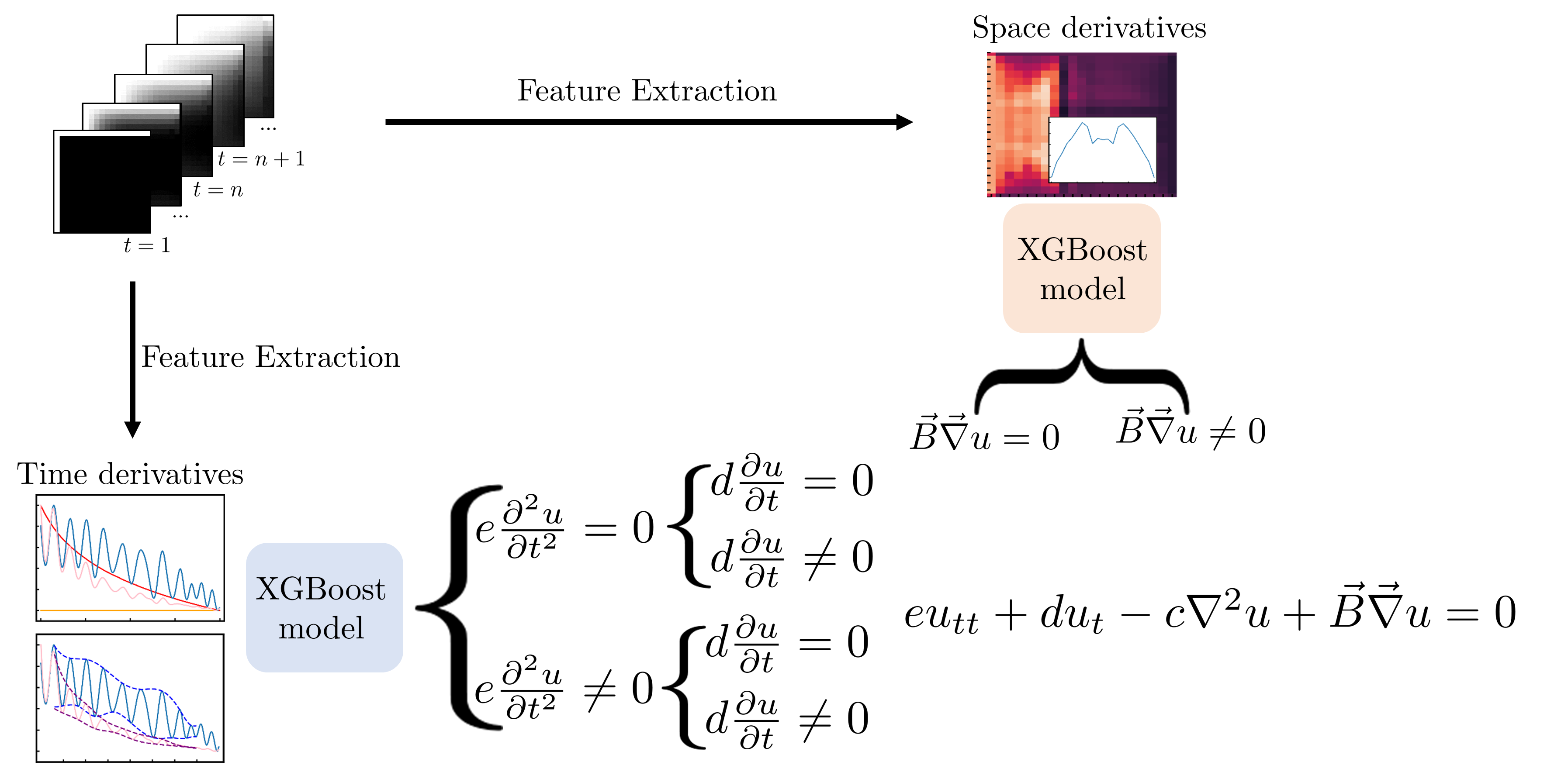}
\end{center}
  \caption{Schematic flowchart of the methodology process. Suitable features are extracted to detect presence of terms in the general form. By adding building blocks together, the equation is identified.}
\label{fig:long}
\end{figure*}
    
    Ensemble methods like XGBoost are highly interpretable \cite{interpretability} which is significantly useful for feature evaluation. In decision tree models, the information gain in the model can be used to find the relative importance of features based on their contribution in prediction. The scores are normalized on the number of detailed features used in each feature type and relative score percentages are calculated. A primary analysis of the models shows that while features have relative importance for classification of data based on PDEs, some expected features are of particular importance for the detection of time or space derivatives (Fig. 5).
    
\begin{figure*}[!htpb]    
\begin{center}
  \includegraphics[width=1\linewidth]{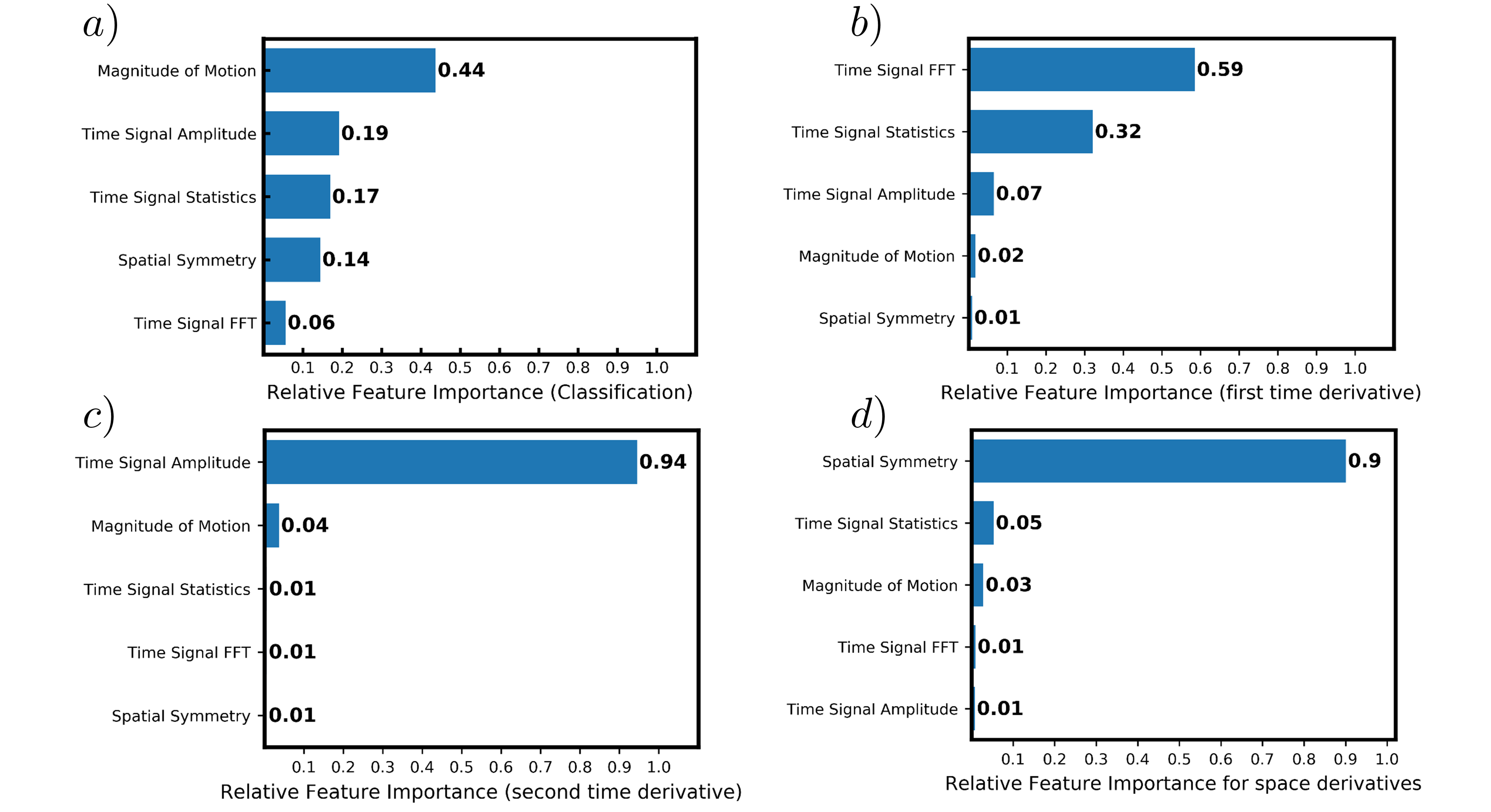}
\end{center}
  \caption{ Relative Feature importance based on information gain in the fitted XGBoost model on data. \textbf{a)} Feature importance for classification of various PDEs without a detailed view of terms. Magnitude of motion is the most effective feature in this task. \textbf{b)} Feature importance to detect presence of the second time derivative ($u_{tt}$) trained on all classes. \textbf{c)} Detection of first time derivative ($u_t$) among all classes. The feature of frequency analysis (FFT) is the most important feature for this task. \textbf{d)} Space derivative ($\nabla u$), the symmetry of motion is the prominent feature for this task.}
\label{fig:long}
\end{figure*}

\section{Results}

\subsection{Data}
  We use numerical simulation to solve PDEs in a two-dimensional space domain. COMSOL multiphysics \cite{COMSOL-Multiphysics} implements Finite Element Method (FEM) to generate the data by defining initial and boundary conditions as well as coefficients. Figure 6a depicts the meshed square domain with Dirichlet boundary conditions on its four sides. The data output, however, is in regular grids of size $21\times 21$ spatially computed for $500$ time steps. 384 samples are generated for 8 classes of PDEs and solutions are normalized by scaling into $[0,1]$ (Fig. 6b).   
  The parameters used for data samples are summarized in \textcolor{blue}{SI table S5}.
  
\begin{figure*}[!htpb]    
\begin{center}
  \includegraphics[width=0.8\linewidth]{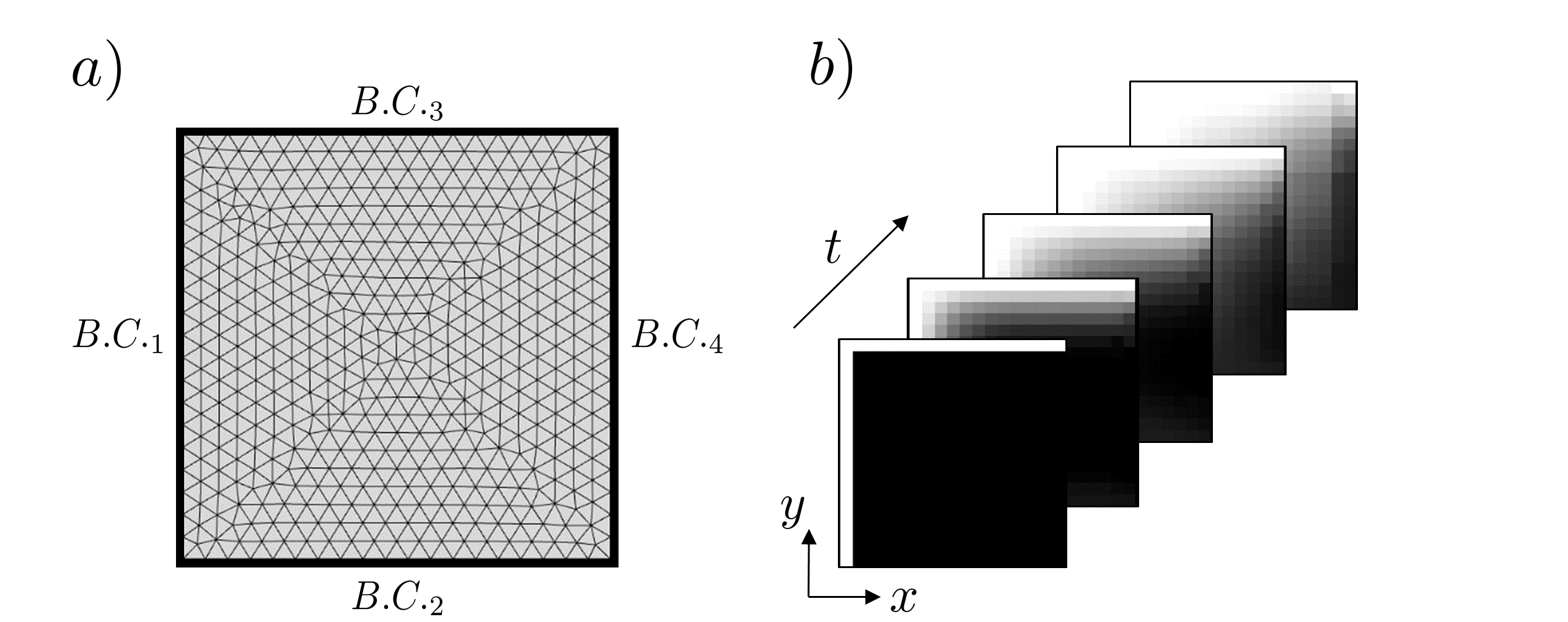}
\end{center}
  \caption{ \textbf{a)} Mesh and Boundary Conditions for data generation using Finite Element Method \textbf{b)} The spatiotemporal data shape for each sample.}
\label{fig:long}
\end{figure*}

\subsection{Experiments}
  As discussed in the previous section, we can use XGBoost trained models to predict terms for a new observation and identify the equation. Here, we design two experiments to evaluate the model. First, we split the data randomly into train/test sets following an 80/20 ratio. Therefore, the model is trained on all the PDEs in the dataset, and the test set has the same statistical distribution of the training set. However, we cannot discover unseen equations with this model. As the second experiment, we can leave the data samples of one specific equation out of training set, so the model is fit on the data from the remaining equations. Next, the samples of the unseen equation are passed to the test section. The results of this experiment would indicate the ability of the model to learn the background physics and discover new equations. To evaluate the features, we compare our model with a Deep Learning (DL) model, given the same amounts of data. We also perform an ablation study to see the effect of features. 

\subsubsection{Automatic Feature Detection}
    Convolutional Neural Networks (CNNs) have made revolutions in many engineering fields due to their significant properties. One of the main advantages of using convolutions is to automatically extract features that contribute to the learning process. Two-dimensional CNNs are designed specifically for images and 2D data. For spatiotemporal data, spatially 2D, we can make use of 3D CNNs where convolution filters are in three dimensions. We trained a CNN on raw data samples which can detect some features and utilize them for the prediction. 
    
    For the good of network, we first preprocess the data samples to the size of $21\times21\times21$ by selecting equally-distanced time steps from the original data. Figure 7 depicts the architecture of the network used in this work, and some of the network and its training parameters are summarized in \textcolor{blue}{SI Table S6}. We used Pytorch deep learning framework to implement the model \cite{pytorch}.
    
\begin{figure*}[!htpb]    
\begin{center}
  \includegraphics[width=1\linewidth]{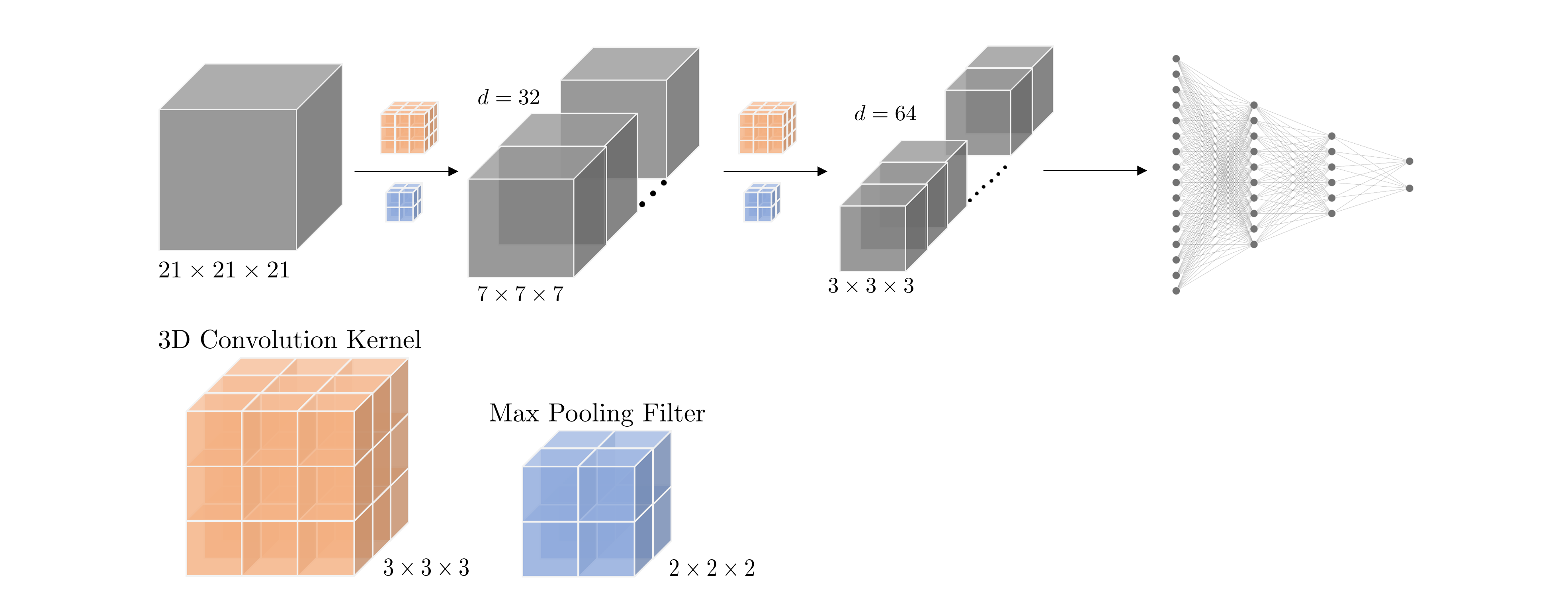}
\end{center}
  \caption{Schematic network architecture of the Three-dimensional Convolutional Neural Network used for classification based on automatic feature detection. Detected features after two layers of convolution are passed through fully connected layers with Softmax classifier.  The convolution kernel and max pooling filter size are also shown.}
\label{fig:long}
\end{figure*}

 \subsubsection{Ablation Study}
  To evaluate the effect of extracted features on the performance of the model, we set up an ablation study in which we perform the classification task (80/20 train/test split ratio) using different sets of features. The results of this study indicate the significance of features to obtain a highly accurate model (Fig. 8a). 
 
    We should notice that in order to get sound results for each term, we should feed the model with the tailored and related features for the very task. For example, if the feature of time dependent signal amplitude is robust for classification of time derivatives, it is meanwhile informationless for spatial derivatives. The same claim holds true for the feature of motion direction symmetry that while is powerful in distinguishing spatial derivatives, it cannot help for time derivatives.
    
 Robust features, i.e. features with meaningful physical background, result in the transferability of the model to new observations on which the model was not trained. While both XGBoost and 3D CNN models perform almost perfect for supervised classification of PDEs, the detected features in 3D CNN are not robust enough to be transferred to new data \cite{transfer-learning}, which means failure in the discovery of new PDEs. However, using the aforementioned physical features, we can predict new equations with high reliability (Fig. 8b). Table 2 shows the performance of the model in the discovery of different equations. In each experiment, the model was trained on seven equations, except one PDE, and the predictions are the results of test on the remaining equation.

\begin{figure*}[!htpb]    
\begin{center}
  \includegraphics[width=1\linewidth]{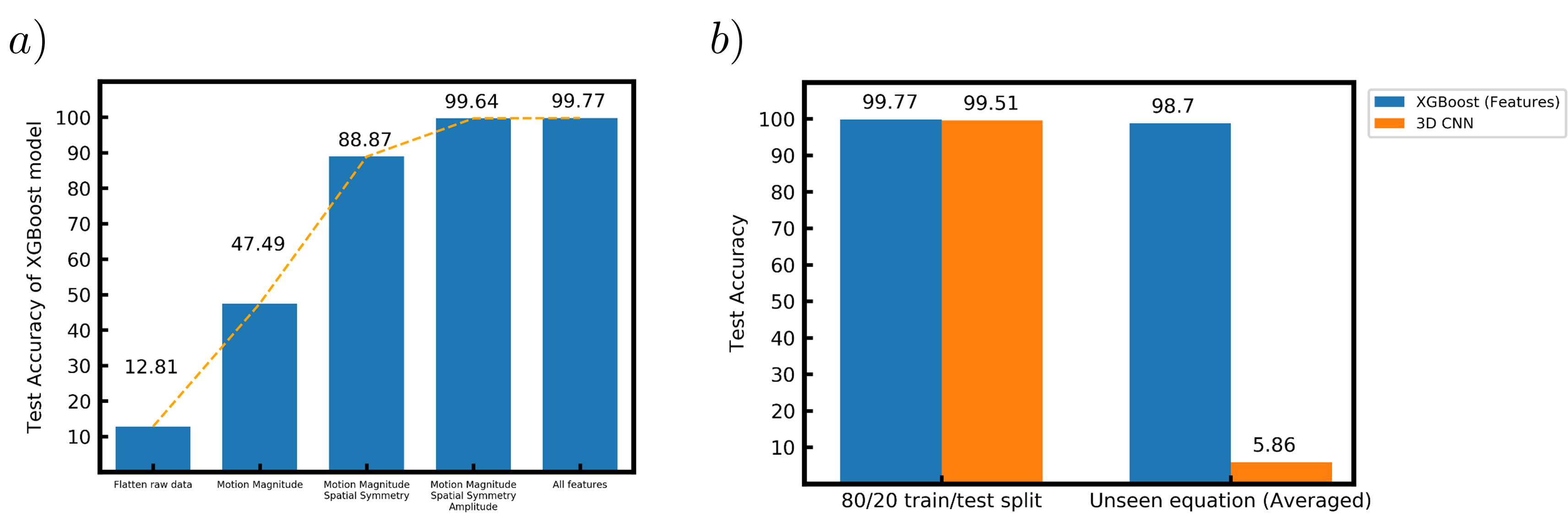}
\end{center}
  \caption{ \textbf{a)} Ablation study on the features used in the XGBoost model; Extracted features result in high test accuracy. \textbf{b)} Comparison of the XGBoost model using extracted features with the 3D Convolutional Neural Network results trained for 100 epochs. Results indicate that while the 3D CNN model has high test accuracy when training and test data are from the same distribution, it fails to identify left-out equations. Robust features, on the other hand, lead to high averaged accuracy in PDE discovery.}
\label{fig:long}
\end{figure*}

    \renewcommand{\arraystretch}{1.3}
    \begin{table}[htbp]
    \begin{center}
    \caption{Test predictions for unseen equations}
    \resizebox{\textwidth}{!}{\begin{tabular}{|c|c|c|c|c|c|c|c|c|c|c|}    
    \hline
    Equation & \diagbox[width=10em]{Ground\\ Truth}{Prediction} &  $1$  &$2$ &$3$ &$4$ & $5$&$6$ & $7$ & $8$& Accuracy   \\
    \hline
    $u_{t} - c \nabla^{2}u  = 0$
     & $1$ &  382 & - & - & - & - & - & \textcolor{red}{2} & - & 99.48  \\
     \hline
     $u_{t} - c \nabla^{2}u + B\nabla u = 0$
     & $2$ & \textcolor{red}{1} & 383 & - & - & - & - & - & - & 99.74  \\
     \hline
     $- c \nabla^{2}u = 0$
     & $3$ & - & - & 384 & - & - & - & - & - & 100.0 \\
     \hline
     $ - c \nabla^{2}u + B\nabla u = 0$
     & $4$ & - & - & \textcolor{red}{5} & 379 & - & - & - & - & 98.70 \\
     \hline
     $u_{tt}  - c \nabla^{2}u  = 0$
     & $5$ & - & - & - & -& 363 & \textcolor{red}{18} & \textcolor{red}{3}  & - & 94.53  \\
     \hline
     $u_{tt} - c \nabla^{2}u + B\nabla u = 0$
     & $6$ & - & - & - & - & - & 384 & - & - & 100.0  \\
     \hline
     $u_{tt} + du_{t} - c \nabla^{2}u = 0$
     & $7$ & - & - & - & - & \textcolor{red}{2} & - & 382 & - & 99.48  \\
     \hline
     $u_{tt} + du_{t} - c \nabla^{2}u + B\nabla u = 0$
     & $8$ & - & - & - & - & - &\textcolor{red}{5} & \textcolor{red}{4} & 375 & 97.66  \\
    \hline
    \multicolumn{11}{r}{Average Accuracy \quad \quad 98.70\% }
    \end{tabular}}
    \end{center}
    \end{table}
  
\subsubsection{Coefficients}
 The extent to which a physical behavior exists in the system and dominates the field usually depends on the magnitude of the coefficients. In fact, we could identify the presence of terms in the last section, however, the coefficients are essentially continuous. Again, based on the gained insight into the meaning of terms, we can compare and compute the magnitude of coefficients for various terms. For example, coefficient $c$ in $-c\nabla^2 u$ corresponds to the wave propagation speed, and thus by detecting the speed of a point like the leading edge of the wave, we can have a measure of this coefficient (Fig. 9a). As another example, the coefficient $d$ in $du_t$ in the damped wave equation corresponds to the extent to which the wave is damped over time which was featured by the decay in the time signal amplitude (Fig. 9b).
 Also Note that given an appropriate structure of data and the known form of equation, the coefficients can be computed by simple regression frameworks (without the need for proposed dictionary of functional terms \cite{PDE-Find})
 
\begin{figure*}[!htpb]    
\begin{center}
  \includegraphics[width=1\linewidth]{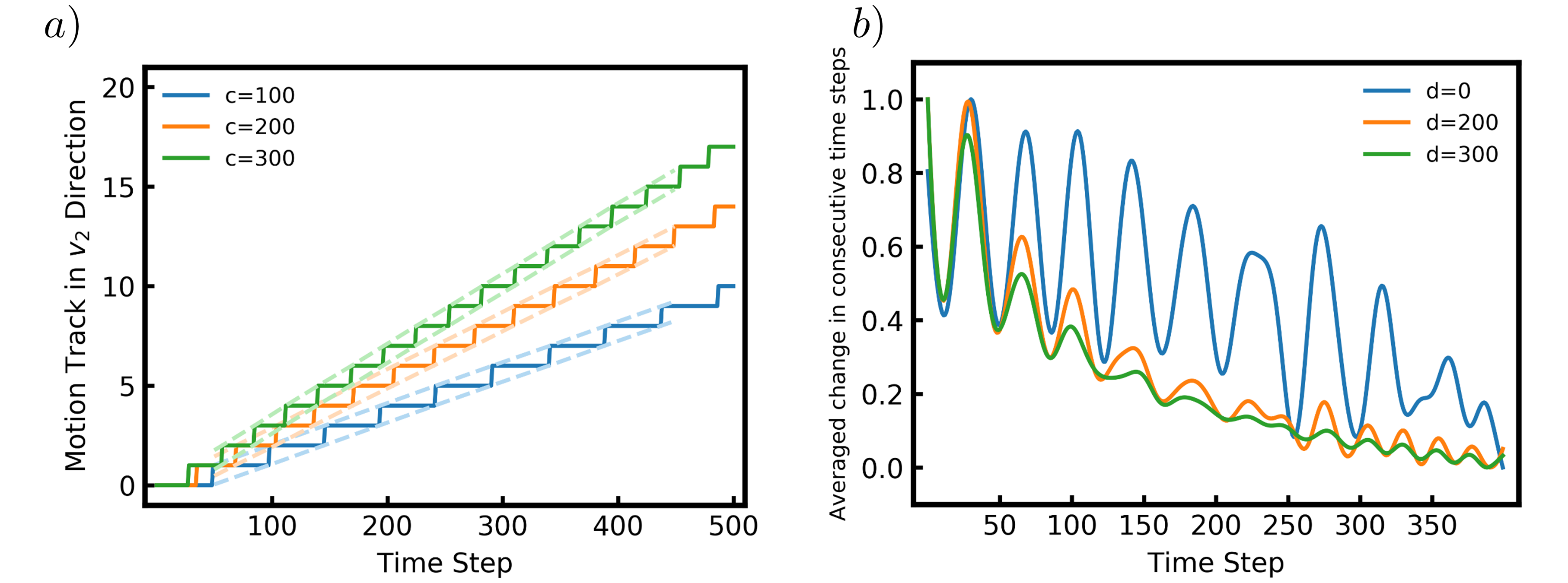}
\end{center}
  \caption{ \textbf{a)} Motion tracking of the wave front edge with envelopes in direction $v_2$ (Eq. 5b). Coefficient $c$ represents velocity and correlates with the slope of envelopes. \textbf{b)} Time signal of change in the system for various magnitudes of damping factor. Higher damping coefficient results in a more damped amplitude of the signal.}
\label{fig:long}
\end{figure*}
 
\section{Conclusion and Discussion}

We have presented a data-driven method to identify Partial Differential Equations governing spatiotemporal data. We leveraged machine learning algorithms as well as prior physical intuition on the data to discover the equations based on useful extracted features. Incorporating this knowledge facilitates the identification process and leads to a robust model that does not rely on differentiation methods. To the best of our knowledge, the proposed method is the first approach that combines physical features with machine learning algorithms for automatic identification of underlying mathematical models. The framework proposed in this paper, however, is limited by the PDEs and the extent of our current knowledge on the physical interpretation of mathematical models.  Therefore, the idea needs further development to be applied to various phenomena. 

Another possible issue is the generalizability of the models. Machine learning models are known to have failures when extrapolating, which means their inability to correctly predict for data samples that are far from the training data distribution. In this work, normalization of data, as well as feature extraction methods, brought about robust features that resulted in highly accurate transferability of the model to the unobserved data, that obviously are out of distribution. However, this issue requires further investigation, especially when applied in extreme cases.

\section*{Declaration of competing interest}
The authors declare that they have no known competing financial interests or personal relationships that could have appeared to influence the work reported in this paper.

\section*{Acknowledgments}
This work is supported by the start-up fund provided by CMU Mechanical Engineering and funding from Sandia National Laboratories.

\bibliography{main}

\begin{thebibliography}{10}

\bibitem{solvePDE-Han}
J.~Han, A.~Jentzen, W.~E, Solving high-dimensional partial differential
  equations using deep learning.
\newblock {\it Proceedings of the National Academy of Sciences\/} {\bf 115},
  8505--8510 (2018).

\bibitem{DGM}
J.~Sirignano, K.~Spiliopoulos, Dgm: A deep learning algorithm for solving
  partial differential equations.
\newblock {\it Journal of Computational Physics\/} {\bf 375}, 1339 - 1364
  (2018).

\bibitem{deepxde}
L.~Lu, X.~Meng, Z.~Mao, G.~E. Karniadakis, Deepxde: A deep learning library for
  solving differential equations (2019).

\bibitem{wiewel2018latentspace}
S.~Wiewel, M.~Becher, N.~Thuerey, Latent-space physics: Towards learning the
  temporal evolution of fluid flow (2018).

\bibitem{farimani2017deep}
A.~B. Farimani, J.~Gomes, V.~S. Pande, Deep learning the physics of transport
  phenomena (2017).

\bibitem{LSTMchaos}
P.~R. Vlachas, W.~Byeon, Z.~Y. Wan, T.~P. Sapsis, P.~Koumoutsakos, Data-driven
  forecasting of high-dimensional chaotic systems with long short-term memory
  networks.
\newblock {\it Proceedings of the Royal Society A: Mathematical, Physical and
  Engineering Sciences\/} {\bf 474}, 20170844 (2018).

\bibitem{kim_lee_2020}
J.~Kim, C.~Lee, Prediction of turbulent heat transfer using convolutional
  neural networks.
\newblock {\it Journal of Fluid Mechanics\/} {\bf 882}, A18 (2020).

\bibitem{PINN-inverse}
M.~{Raissi}, P.~{Perdikaris}, G.~E. {Karniadakis}, {Physics-informed neural
  networks: A deep learning framework for solving forward and inverse problems
  involving nonlinear partial differential equations}.
\newblock {\it Journal of Computational Physics\/} {\bf 378}, 686-707 (2019).

\bibitem{CPINN}
A.~D. Jagtap, E.~Kharazmi, G.~E. Karniadakis, Conservative physics-informed
  neural networks on discrete domains for conservation laws: Applications to
  forward and inverse problems.
\newblock {\it Computer Methods in Applied Mechanics and Engineering\/} {\bf
  365}, 113028 (2020).

\bibitem{DL_PhysicalProcesses}
E.~de~B{\'{e}}zenac, A.~Pajot, P.~Gallinari, Deep learning for physical
  processes: incorporating prior scientific knowledge.
\newblock {\it Journal of Statistical Mechanics: Theory and Experiment\/} {\bf
  2019}, 124009 (2019).

\bibitem{PI-deep-generative}
Y.~Yang, P.~Perdikaris, Physics-informed deep generative models (2018).

\bibitem{weaklysupervised-Barati}
R.~Sharma, A.~B. Farimani, J.~Gomes, P.~Eastman, V.~Pande, Weakly-supervised
  deep learning of heat transport via physics informed loss (2018).

\bibitem{hamiltonian}
S.~Greydanus, M.~Dzamba, J.~Yosinski, {\it Advances in Neural Information
  Processing Systems 32\/}, H.~Wallach, H.~Larochelle, A.~Beygelzimer,
  F.~d\textquotesingle Alch\'{e}-Buc, E.~Fox, R.~Garnett, eds. (Curran
  Associates, Inc., 2019), pp. 15379--15389.

\bibitem{Crutchfield1987}
J.~P. Crutchfield, B.~S. McNamara, Equations of motion from a data series.
\newblock {\it Complex Systems\/} {\bf 1} (1987).

\bibitem{Bongard-Lipson}
J.~Bongard, H.~Lipson, Automated reverse engineering of nonlinear dynamical
  systems.
\newblock {\it Proceedings of the National Academy of Sciences\/} {\bf 104},
  9943--9948 (2007).

\bibitem{Schmidt-Lipson}
M.~Schmidt, H.~Lipson, Distilling free-form natural laws from experimental
  data.
\newblock {\it Science\/} {\bf 324}, 81--85 (2009).

\bibitem{SINDY}
S.~L. Brunton, J.~L. Proctor, J.~N. Kutz, Discovering governing equations from
  data by sparse identification of nonlinear dynamical systems.
\newblock {\it Proceedings of the National Academy of Sciences\/} {\bf 113},
  3932--3937 (2016).

\bibitem{PDE-Find}
S.~H. Rudy, S.~L. Brunton, J.~L. Proctor, J.~N. Kutz, Data-driven discovery of
  partial differential equations.
\newblock {\it Science Advances\/} {\bf 3} (2017).

\bibitem{Schaeffer}
H.~Schaeffer, Learning partial differential equations via data discovery and
  sparse optimization.
\newblock {\it Proceedings of the Royal Society A: Mathematical, Physical and
  Engineering Sciences\/} {\bf 473}, 20160446 (2017).

\bibitem{Parametric-PDE}
S.~Rudy, A.~Alla, S.~L. Brunton, J.~N. Kutz, Data-driven identification of
  parametric partial differential equations.
\newblock {\it SIAM Journal on Applied Dynamical Systems\/} {\bf 18}, 643-660
  (2019).

\bibitem{Chang-2019}
H.~Chang, D.~Zhang, Identification of physical processes via combined
  data-driven and data-assimilation methods.
\newblock {\it Journal of Computational Physics\/} {\bf 393}, 337 - 350 (2019).

\bibitem{raissi-Gaussian}
M.~Raissi, P.~Perdikaris, G.~E. Karniadakis, Machine learning of linear
  differential equations using gaussian processes.
\newblock {\it J. Comput. Phys.\/} {\bf 348}, 683–693 (2017).

\bibitem{hidden-physics}
M.~Raissi, G.~E. Karniadakis, Hidden physics models: Machine learning of
  nonlinear partial differential equations.
\newblock {\it Journal of Computational Physics\/} {\bf 357}, 125 - 141 (2018).

\bibitem{universal-f-apprx}
K.~Hornik, M.~Stinchcombe, H.~White, Multilayer feedforward networks are
  universal approximators.
\newblock {\it Neural Networks\/} {\bf 2}, 359 - 366 (1989).

\bibitem{discovery-complex}
J.~Berg, K.~Nyström, Data-driven discovery of pdes in complex datasets.
\newblock {\it Journal of Computational Physics\/} {\bf 384}, 239 - 252 (2019).

\bibitem{dl-pde}
H.~Xu, H.~Chang, D.~Zhang, Dl-pde: Deep-learning based data-driven discovery of
  partial differential equations from discrete and noisy data (2019).

\bibitem{raissi-multistep}
M.~Raissi, P.~Perdikaris, G.~E. Karniadakis, Multistep neural networks for
  data-driven discovery of nonlinear dynamical systems (2018).

\bibitem{DeepHiddenPhysics}
M.~Raissi, Deep hidden physics models: Deep learning of nonlinear partial
  differential equations.
\newblock {\it Journal of Machine Learning Research\/} {\bf 19}, 1-24 (2018).

\bibitem{PDE-Net}
Z.~Long, Y.~Lu, X.~Ma, B.~Dong, {\it Proceedings of the 35th International
  Conference on Machine Learning\/}, J.~Dy, A.~Krause, eds. (PMLR,
  Stockholmsmässan, Stockholm Sweden, 2018), vol.~80 of {\it Proceedings of
  Machine Learning Research\/}, pp. 3208--3216.

\bibitem{PDE-Net-2.0}
Z.~Long, Y.~Lu, B.~Dong, Pde-net 2.0: Learning pdes from data with a
  numeric-symbolic hybrid deep network.
\newblock {\it Journal of Computational Physics\/} {\bf 399}, 108925 (2019).

\bibitem{goodfellow2014-adversarial}
I.~J. Goodfellow, J.~Shlens, C.~Szegedy, Explaining and harnessing adversarial
  examples (2014).

\bibitem{adversarial-NIPS2019}
A.~Ilyas, S.~Santurkar, D.~Tsipras, L.~Engstrom, B.~Tran, A.~Madry, {\it
  Advances in Neural Information Processing Systems 32\/}, H.~Wallach,
  H.~Larochelle, A.~Beygelzimer, F.~d\textquotesingle Alch\'{e}-Buc, E.~Fox,
  R.~Garnett, eds. (Curran Associates, Inc., 2019), pp. 125--136.

\bibitem{transferability-adv}
N.~Papernot, P.~McDaniel, I.~Goodfellow, Transferability in machine learning:
  from phenomena to black-box attacks using adversarial samples (2016).

\bibitem{savitzky}
A.~Savitzky, M.~J.~E. Golay, Smoothing and differentiation of data by
  simplified least squares procedures.
\newblock {\it Analytical Chemistry\/} {\bf 36}, 1627-1639 (1964).

\bibitem{motion-detection}
J.~Hernandez, H.~Morita, M.~Nakano-Miytake, H.~Perez-Meana, {\it Progress in
  Pattern Recognition, Image Analysis, Computer Vision, and Applications\/},
  E.~Bayro-Corrochano, J.-O. Eklundh, eds. (Springer Berlin Heidelberg, Berlin,
  Heidelberg, 2009), pp. 1054--1061.

\bibitem{xgboost}
T.~Chen, C.~Guestrin, {\it Proceedings of the 22nd ACM SIGKDD International
  Conference on Knowledge Discovery and Data Mining\/}, KDD ’16 (Association
  for Computing Machinery, New York, NY, USA, 2016), p. 785–794.

\bibitem{scikit-learn}
F.~Pedregosa, G.~Varoquaux, A.~Gramfort, V.~Michel, B.~Thirion, O.~Grisel,
  M.~Blondel, P.~Prettenhofer, R.~Weiss, V.~Dubourg, J.~Vanderplas, A.~Passos,
  D.~Cournapeau, M.~Brucher, M.~Perrot, E.~Duchesnay, Scikit-learn: Machine
  learning in {P}ython.
\newblock {\it Journal of Machine Learning Research\/} {\bf 12}, 2825--2830
  (2011).

\bibitem{interpretability}
Z.~C. Lipton, The mythos of model interpretability.
\newblock {\it Queue\/} {\bf 16}, 31–57 (2018).

\bibitem{COMSOL-Multiphysics}
C.~Multiphysics., v5.3a.

\bibitem{pytorch}
A.~Paszke, S.~Gross, F.~Massa, A.~Lerer, J.~Bradbury, G.~Chanan, T.~Killeen,
  Z.~Lin, N.~Gimelshein, L.~Antiga, A.~Desmaison, A.~Kopf, E.~Yang, Z.~DeVito,
  M.~Raison, A.~Tejani, S.~Chilamkurthy, B.~Steiner, L.~Fang, J.~Bai,
  S.~Chintala, {\it Advances in Neural Information Processing Systems 32\/},
  H.~Wallach, H.~Larochelle, A.~Beygelzimer, F.~d\textquotesingle
  Alch\'{e}-Buc, E.~Fox, R.~Garnett, eds. (Curran Associates, Inc., 2019), pp.
  8024--8035.

\bibitem{transfer-learning}
J.~Yosinski, J.~Clune, Y.~Bengio, H.~Lipson, {\it Proceedings of the 27th
  International Conference on Neural Information Processing Systems - Volume
  2\/}, NIPS’14 (MIT Press, Cambridge, MA, USA, 2014), p. 3320–3328.

\end{thebibliography}


\begin{thebibliography}{1}

\bibitem{savitzky}
A.~Savitzky, M.~J.~E. Golay, Smoothing and differentiation of data by
  simplified least squares procedures.
\newblock {\it Analytical Chemistry\/} {\bf 36}, 1627-1639 (1964).

\bibitem{COMSOL-Multiphysics}
C.~Multiphysics., v5.3a.

\end{thebibliography}
\end{document}


\begin{frontmatter}
	
\title{Supplementary Information for: \\ 
Data-driven Identification of 2D Partial Differential Equations using extracted physical features
}
    
\author[1]{Kazem Meidani}
\author[1,2,3]{Amir Barati Farimani\corref{main}}\cortext[main]{Corresponding author}
\ead{barati@cmu.edu}
    
\address[1]{Department of Mechanical Engineering\unskip, 
    Carnegie Mellon University, Pittsburgh, PA, USA}
\address[2]{Department of Chemical Engineering\unskip, 
    Carnegie Mellon University, Pittsburgh, PA, USA}
\address[3]{Machine Learning Department\unskip, 
Carnegie Mellon University, Pittsburgh, PA, USA}

\end{frontmatter}

\section{Feature Extraction}

\subsection{Time derivatives}
  
  Savitzky-Golay filter \cite{savitzky} uses convolutions to smoothen the sections of signal. In our case, we use 3rd order of polynomial fitting on a window length of size 21 that strides over time steps. Figure S1 illustrates a sample of smoothed diffusion signal compared with its raw signal.
  
\begin{figure*}[!htpb]    
\begin{center}
  \includegraphics[width=0.8\linewidth]{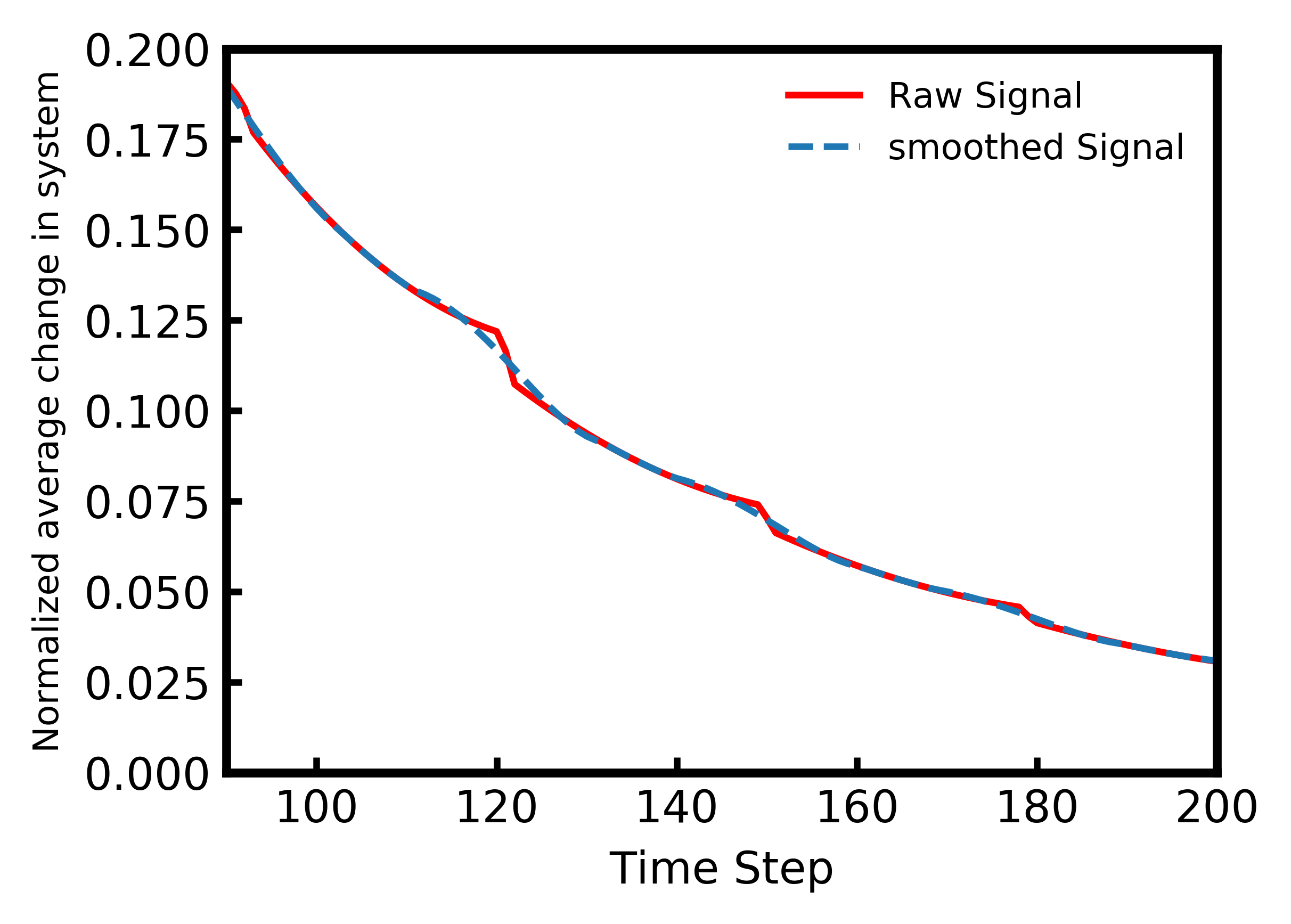}
\end{center}
  \caption{Raw time signal for a diffusion sample vs. the smoothed signal using Savitzky-Golay filter.}
\label{fig:long}
\end{figure*}
  
   \subsection{Statistics of the signal} 
   
   Table S1 contains the details of features used as statistical characteristics of the time signal of averaged temporal change in consecutive time steps. 
   
    \renewcommand{\arraystretch}{1.3}
    \begin{table}[htbp]
    \begin{center}
    \caption{List of features from the statistics of the extracted time signal (Averaged change in every consecutive time steps)}
    {\begin{tabular}{ll|ll}
    \hline
    \multicolumn{4}{c}{$\mathcal{F}_{statistics}$ }             \\
    \hline
    $\mathcal{F}_{1}$ &   $Mean (u)|^{t=450}_{t=50}$  &
    $\mathcal{F}_{2}$   & $Std (u)|^{t=450}_{t=50}$                  \\
    $\mathcal{F}_{3}$  & $Min (u)|^{t=450}_{t=350}$  &
    $\mathcal{F}_{4}$ & $Max (u)|^{t=450}_{t=350}$      \\ 
    $\mathcal{F}_{5}$    & $Mean (u)|^{t=450}_{t=350}$  &
    $\mathcal{F}_{6}$ &$Std (u)|^{t=450}_{t=350}$  \\ 
    $\mathcal{F}_{7}$ & $Skew (u)|^{t=450}_{t=50}$   &  &   \\ 

    \hline
    \end{tabular}}
    \end{center}
    \footnotesize{Std: Standard Deviation, Skew: Skewness, T=500}
    \end{table}

   \subsection{Oscillations and Amplitude of the signal}  
   
    Upper and lower envelopes of the aforementioned signals are computed with interpolation of effective peaks and valleys of the signal (Equation S1a). So, the amplitude of the signal would be the difference between these envelopes (Equation S1b). Fig S2 depicts the signals and their amplitude for wave and damped wave equations. 

    \begin{subequations}
    \begin{equation}
    (Upper/Lower)\, Envelope = Cubic \, Interpolation \,  (peaks/valleys)
    \end{equation}
    \begin{equation}
    A = Upper Envelope - Lower Envelope
    \end{equation} 
    \end{subequations}
    
\begin{figure*}[!htpb]    
\begin{center}
  \includegraphics[width=0.8\linewidth]{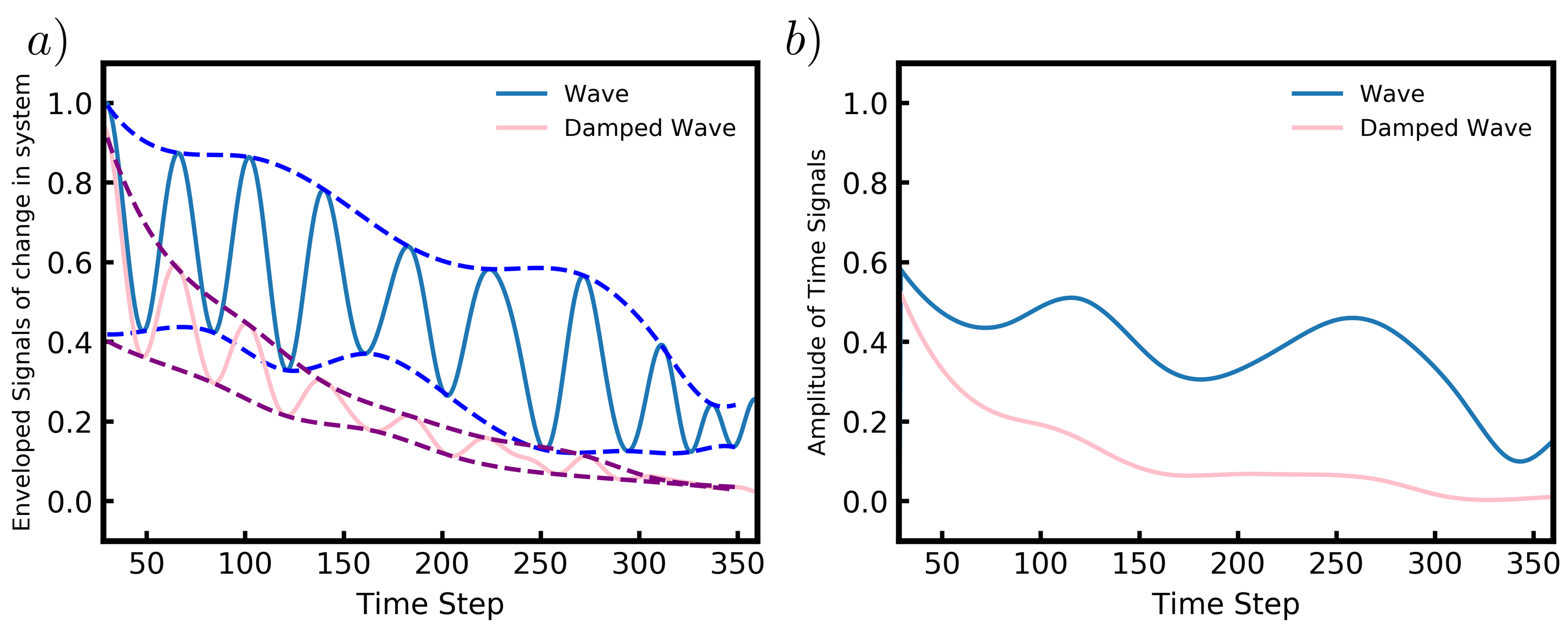}
\end{center}
  \caption{\textbf{a) }Time signals of averaged change in consecutive time steps for two samples of wave and damped wave PDEs. \textbf{b) } Amplitude signal (A) corresponding to these time signals.}
\label{fig:long}
\end{figure*}

      \renewcommand{\arraystretch}{1.3}
    \begin{table}[htbp]
    \begin{center}
    \caption{List of features from the extracted time signal (Averaged change in every consecutive time steps)}
    {\begin{tabular}{ll|ll}
    \hline
    \multicolumn{4}{c}{$\mathcal{F}_{Amplitude}$ }             \\
    \hline
    $\mathcal{F}_{1}$ &   $Max (A)|^{t=450}_{t=50}$& $\mathcal{F}_{8}$ & Number of peaks\\
    $\mathcal{F}_{2}$   & $Arg Max (A)|^{t=450}_{t=50}$ &  $\mathcal{F}_{9}$ &  Number of valleys  \\
    $\mathcal{F}_{3}$  & $Mean(A)|^{t=450}_{t=350}$ & 
    \multirow{2}{*}{$\mathcal{F}_{10}$} &  \multirow{2}{*}{${Mean (A)|^{t=450}_{t=250}}\, / \, {Mean (A)|^{t=250}_{t=50}}$} \\ 
    $\mathcal{F}_{4}$ & $Std (A)|^{t=450}_{t=350}$ &   & \\ 
    $\mathcal{F}_{5}$  & $Skew (A)|^{t=450}_{t=350}$ & 
    \multirow{2}{*}{$\mathcal{F}_{11}$} & \multirow{2}{*}{${Max (A)|^{t=450}_{t=250}} \, / \, {Max (A)|^{t=250}_{t=50}}$ } \\ 
    $\mathcal{F}_{6}$ &  $Mean (A)|^{t=450}_{t=250}$ &   &  \\ 
    $\mathcal{F}_{7}$ &  $Std (A)|^{t=450}_{t=250}$ &   & \\ 
    \hline
    \end{tabular}}
    \end{center}
    \footnotesize{Std: Standard Deviation, Skew: Skewness, T=500}
    \end{table}
   
   \subsection{Frequency analysis of the signal} 
    
    In order to extract features from Fast Fourier Transforms of signals, we divide the main frequency domain ($f <= 0.1$) to 5 bins (Fig. S3). Subsequently, according to table S3, some statistical measures of FFT signal in each of these bins are extracted.
    
\begin{figure*}[!htpb]    
\begin{center}
  \includegraphics[width=0.8\linewidth]{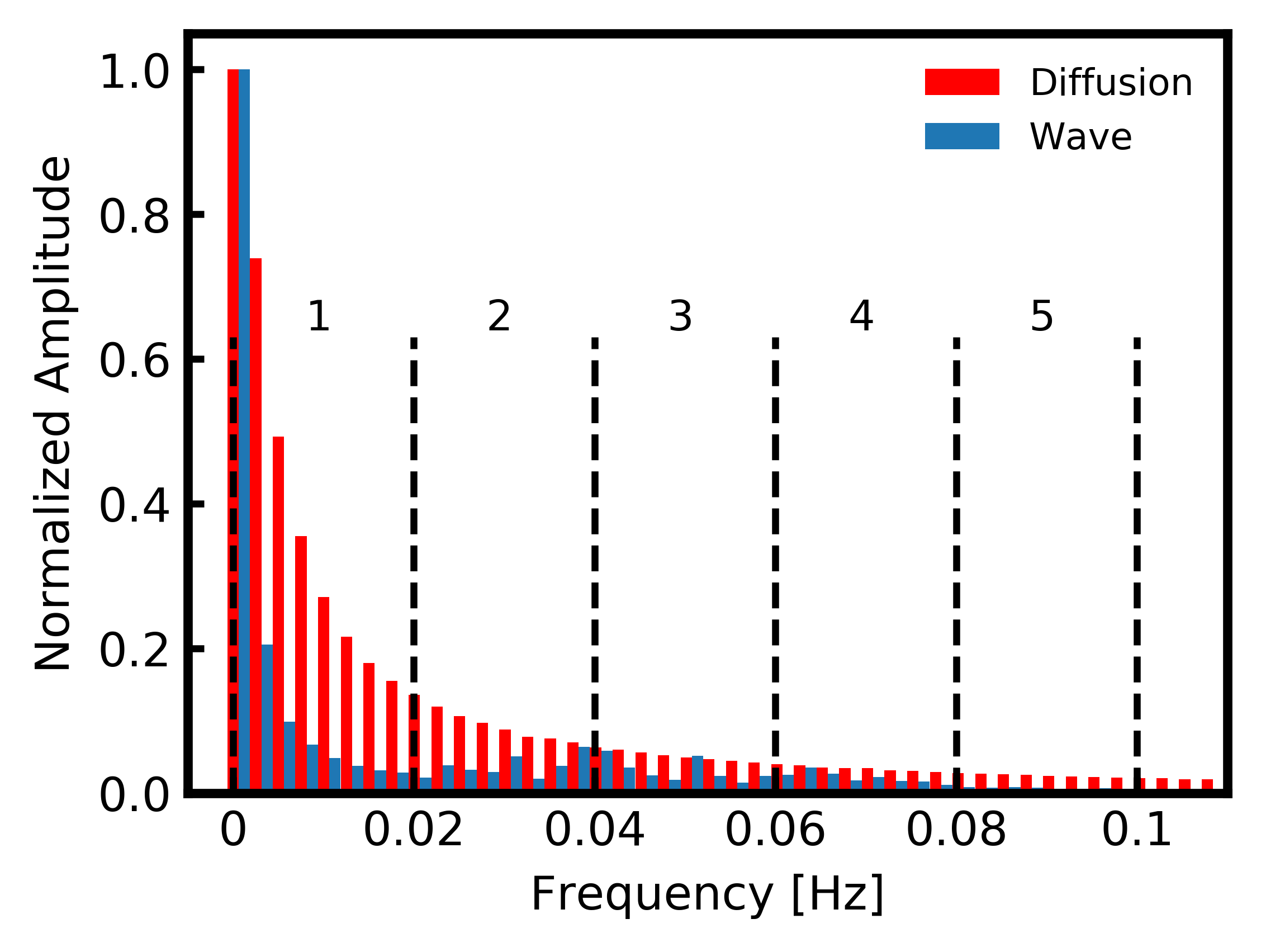}
\end{center}
  \caption{Fast Fourier Transform feature extraction details; Statistics of FFT on 5 equal ranges of frequency are computed as features.}
\label{fig:long}
\end{figure*}

      \renewcommand{\arraystretch}{1.5}
    \begin{table}[htbp]
    \begin{center}
    \caption{List of features from frequency analysis of the time signal (Averaged change in every consecutive time steps)}
     \resizebox{\textwidth}{!}{\begin{tabular}{ll|ll|ll|ll}
    \hline
    \multicolumn{8}{c}{$\mathcal{F}_{FFT}$ }             \\
    \hline
    $\mathcal{F}_{1-1}$ &   $Mean (FFT)|^{f=0.02}_{f=0}$ & $\mathcal{F}_{1-2}$ &   $Std (FFT)|^{f=0.02}_{f=0}$&
    $\mathcal{F}_{1-3}$ &   $Min (FFT)|^{f=0.02}_{f=0}$ & $\mathcal{F}_{1-4}$ &   $Max(FFT)|^{f=0.02}_{f=0}$ \\
    
    $\mathcal{F}_{2-1}$ & $Mean (FFT)|^{f=0.04}_{f=0.02}$ &  $\mathcal{F}_{2-2}$ &   $Std (FFT)|^{f=0.04}_{f=0.02}$ & $\mathcal{F}_{2-3}$ &   $Min (FFT)|^{f=0.04}_{f=0.02}$ &$\mathcal{F}_{2-4}$ &   $Max(FFT)|^{f=0.04}_{f=0.02}$ \\
    
    $\mathcal{F}_{3-1}$  & $Mean (FFT)|^{f=0.06}_{f=0.04}$ & $\mathcal{F}_{3-2}$ &   $Std (FFT)|^{f=0.06}_{f=0.04}$ & 
    $\mathcal{F}_{3-3}$ &   $Min (FFT)|^{f=0.06}_{f=0.04}$ & $\mathcal{F}_{3-4}$ &   $Max(FFT)|^{f=0.06}_{f=0.04}$ \\ 
    
    $\mathcal{F}_{4-1}$ & $Mean (FFT)|^{f=0.08}_{f=0.06}$ &  $\mathcal{F}_{4-2}$ &   $Std (FFT)|^{f=0.08}_{f=0.06}$ & 
    $\mathcal{F}_{4-3}$ &   $Min (FFT)|^{f=0.08}_{f=0.06}$ & $\mathcal{F}_{4-4}$ &   $Max(FFT)|^{f=0.08}_{f=0.06}$ \\
    
    $\mathcal{F}_{5-1}$    & $Mean (FFT)|^{f=0.1}_{f=0.08}$ & $\mathcal{F}_{5-2}$ &   $Std (FFT)|^{f=0.1}_{f=0.08}$&
    $\mathcal{F}_{5-3}$ &   $Min (FFT)|^{f=0.1}_{f=0.08}$ & $\mathcal{F}_{5-4}$ &   $Max(FFT)|^{f=0.1}_{f=0.08}$ \\ 
    \hline
    \end{tabular}}
    \end{center}
    \footnotesize{Std: Standard Deviation, Skew: Skewness, T=500}
    \end{table}

   \subsection{Motion magnitude} 
   
   At each time step t and for every grid point $i,j$, all the data point in a filter of size $3\times3$ around the position of original point are considered at the next time step $t+1$. The motion vector is calculated by finding the position in the filter with minimum error compared to the original grid point (Equation S2).
   
    \begin{subequations}
    \begin{equation}
    MAE(u^t_{i,j}) = \sum_{k=-1}^{1} \sum_{l=-1}^{1} u^{t+1}(i+k,j+l) 
    \end{equation}
    \begin{equation}
    \vec{V} =  MAE(u^t_{i,j}) - u^t_{i,j}
    \end{equation} 
    \end{subequations}

   \subsection{Space derivatives} 
   
   To find the spatial motion feature, we first examine the change of data at one single time step in two main directions $X,Y$ (Equation S3). Based on the gradient of solution intensity, value etc., we can find the diffusion motion direction and its perpendicular vector $\vec{v}_1$(Equation S4). The changes in two main Cartesian directions are projected into these directions. To evaluate the symmetry of the projected vector, the first half of the signal is subtracted from the second part (Equation S7). Table S4 summarizes some of the features used from this signal. Below are two examples of samples with different motion vectors that illustrate the vector and feature extraction process. In the first example (Equation S6, Fig. S4), the motion vector aligns with $Y$ direction but in the second example it is a linear combination of $X,Y$ directions (Equation S7, Fig. S5). 
   
    \begin{equation}
    \bar{u}_{i,x-dir} = \frac{\sum_{j}^{\mathcal{N}_{y}} u^{t}_{i,j}}{\mathcal{N}_{y}}\quad , \quad
    \bar{u}_{j,y-dir} = \frac{\sum_{i}^{\mathcal{N}_{x}} u^{t}_{i,j}}{\mathcal{N}_{x}}
    \end{equation}  

    \begin{subequations}
    \begin{equation}
    \begin{aligned}
    d_1 = \frac{\Delta{\mathcal{C}_x}}{\| \Delta{\mathcal{C}} \|_2} \quad ,\quad
    d_2 = \frac{\Delta{\mathcal{C}_y}} { \| \Delta{\mathcal{C}} \|_2}
    \end{aligned}
    \end{equation}
    \begin{equation}
    \begin{aligned}
    v_1 = d_1 \bar{u}_y - d_2 \bar{u}_x
      \\
    v_2 = d_1 \bar{u}_x + d_2 \bar{u}_y
    \end{aligned}
    \end{equation}
    \end{subequations}
    
    \begin{equation}
    \begin{aligned}
    \mathcal{S} = v_1|^{grid=10}_{grid=0} - v_1|^{gird=10}_{grid=20}
    \end{aligned}
    \end{equation}
    
       \renewcommand{\arraystretch}{1.3}
    \begin{table}[htbp]
    \begin{center}
    \caption{List of features from spatial motion direction}
    {\begin{tabular}{ll|ll}
    \hline
    \multicolumn{4}{c}{$\mathcal{F}_{spatial\, symmetry}$ }             \\
    \hline
    $\mathcal{F}_{1}$ &   $Mean (u)|^{t=450}_{t=50}$  &
    $\mathcal{F}_{2}$   & $Std (u)|^{t=450}_{t=50}$                  \\
    $\mathcal{F}_{3}$  & $Min (u)|^{t=450}_{t=350}$  &
    $\mathcal{F}_{4}$ & $Max (u)|^{t=450}_{t=350}$      \\ 
    $\mathcal{F}_{5}$    & $Mean (u)|^{t=450}_{t=350}$  &
    $\mathcal{F}_{6}$ &$Std (u)|^{t=450}_{t=350}$  \\ 
    $\mathcal{F}_{7}$ & $Skew (u)|^{t=450}_{t=50}$   &  &   \\ 

    \hline
    \end{tabular}}
    \end{center}
    \footnotesize{Std: Standard Deviation, Skew: Skewness, T=500}
    \end{table}

    Example 1)
     \begin{subequations}
     \begin{equation}
    \begin{aligned}
    \Delta{\mathcal{C}_x} = (B.C._{1}-B.C._{4}) = 6
    \quad , \quad 
    \Delta{\mathcal{C}_y}= (B.C._{2}-B.C._{3}) = 0
    \end{aligned}
    \end{equation}
    \begin{equation}
    \begin{aligned}
    d_1 = \frac{\Delta{\mathcal{C}_x}}{\| \Delta{\mathcal{C}} \|_2} = 1 \quad ,\quad
    d_2 = \frac{\Delta{\mathcal{C}_y}} { \| \Delta{\mathcal{C}} \|_2} = 0
    \end{aligned}
    \end{equation}
    \begin{equation}
    \begin{aligned}
    v_1 = d_1 \bar{u}_y - d_2 \bar{u}_x = \bar{u}_y
      \\
    v_2 = d_1 \bar{u}_x + d_2 \bar{u}_y = \bar{u}_x
    \end{aligned}
    \end{equation}
    \end{subequations}

\begin{figure*}[!htpb]    
\begin{center}
  \includegraphics[width=0.9\linewidth]{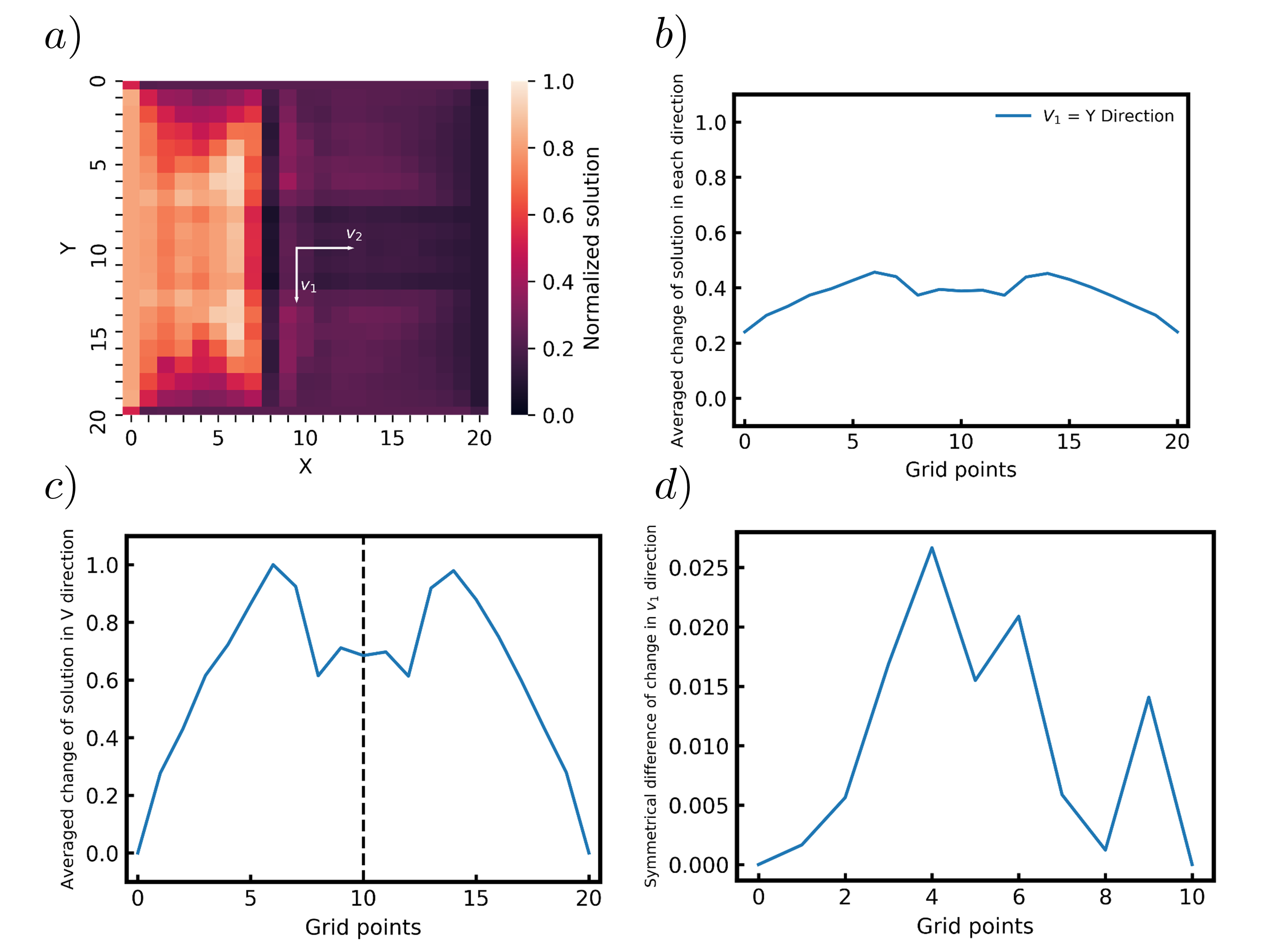}
\end{center}
  \caption{  \textbf{a)} Snapshot of data at time $t$ along with the extracted directions of $\vec{v_1}, \, \vec{v_2}$ \textbf{b)} Changes in the $v_1 = Y$ direction \textbf{c)} Normalized signal of (b)  \textbf{d)} Vector S (Symmetrical difference) that is used for feature extraction.}
\label{fig:long}
\end{figure*}
    
        Example 2)
     \begin{subequations}
     \begin{equation}
    \begin{aligned}
    \Delta{\mathcal{C}_x} = (B.C._{1}-B.C._{4}) = 6
    \quad , \quad 
    \Delta{\mathcal{C}_y}= (B.C._{2}-B.C._{3}) = 5
    \end{aligned}
    \end{equation}
    \begin{equation}
    \begin{aligned}
    d_1 = \frac{\Delta{\mathcal{C}_x}}{\| \Delta{\mathcal{C}} \|_2} = \frac{6}{\sqrt{61}} = 0.768  \quad ,\quad
    d_2 = \frac{\Delta{\mathcal{C}_y}} { \| \Delta{\mathcal{C}} \|_2} =\frac{5}{\sqrt{61}} = 0.640 
    \end{aligned}
    \end{equation}
    \begin{equation}
    \begin{aligned}
    v_1 = 0.768\bar{u}_y - 0.640 \bar{u}_x
      \\
    v_2 = 0.768\bar{u}_x + 0.640 \bar{u}_y 
    \end{aligned}
    \end{equation}
    \end{subequations}
    
\begin{figure*}[!htpb]    
\begin{center}
  \includegraphics[width=0.9\linewidth]{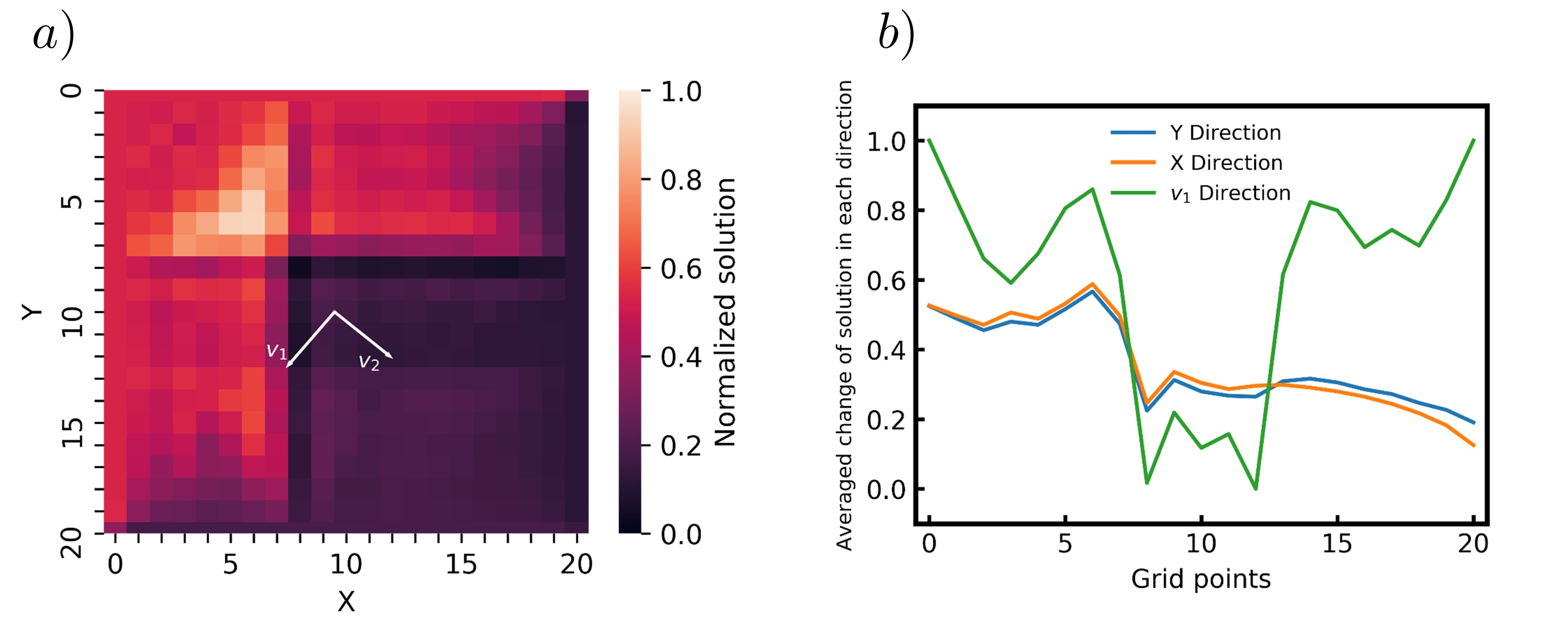}
\end{center}
  \caption{  \textbf{a)} Snapshot of data at time $t$ along with the extracted directions of $\vec{v_1}, \, \vec{v_2}$ \textbf{b)} Changes in the calculated $v_1$ direction. We can see that this direction is perpendicular to the diffusion based motion direction, so preserves approximate symmetry for equations without convection term.}
\label{fig:long}
\end{figure*}

\section{Results}

\subsection{Data}
  The parameters used for data samples are summarized in table S5. The data is numerically simulated by COMSOL \cite{COMSOL-Multiphysics} under similar conditions and normalization (Equation S8).

     \begin{subequations}
     \begin{equation}
    \begin{aligned}
    dt = 1e-4 \, , \quad T=[0,0.05] \, , \quad X,Y \in [21\times21] \, ,\quad I.C.=0.1
    \end{aligned}
    \end{equation}
    \begin{equation}
    \begin{aligned}
    u_N(x,y,t) = \frac{u(x,y,t)-Min(u)}{Max(u)-Min(u)} 
    \end{aligned}
    \end{equation}
    \end{subequations}

  \begin{table}[htbp]
    \begin{center}
    \caption{Equations and the parameters in data generation}
    \resizebox{\textwidth}{!}{\begin{tabular}{|l|l|l|l|l|}    
    \hline
    \multirow{2}{*}{Equation} &  \multicolumn{2}{c}{Coefficients}   &   \multirow{2}{*}{Boundary Conditions}  \\
    &  Time derivatives  & Space derivatives &    \\
    \hline
     \multirow{2}{*}{$u_{t} - c \nabla^{2}u = 0$} 
     & \multirow{2}{*}{$e=0 ,\, d=1$}  & $c=\{1,3,5,7,9,11\}$ & $BC_1 , BC_2, BC_3 $  \\
     &  & $B = [0,0]$ & $\in \{-4,1,6,11\}$  \\
      \hline
     \multirow{2}{*}{$u_{t} - c \nabla^{2}u + B\nabla u = 0$} 
     & \multirow{2}{*}{$e=0 ,\, d=1$}  & $c=\{1,6,11\}$ & $ BC_1 , BC_2, BC_3$   \\
     &  & $B_x , B_y \in \{70,90,110,130\}$ & $\in \{1,6\}$  \\
      \hline
     \multirow{2}{*}{$-c \nabla^{2}u  = 0$} 
     & \multirow{2}{*}{$e=0 ,\, d=1$}  & $c=\{1,3,5,7,9,11\}$ & $BC_1 , BC_2, BC_3  $ \\
     &  & $B = [0,0]$ & $\in \{-4,1,6,11\}$  \\
      \hline
     \multirow{2}{*}{$-c \nabla^{2}u + B\nabla u = 0$} 
     & \multirow{2}{*}{$e=0 ,\, d=1$}  & $c=\{5,8,11\}$ & $BC_2 , BC_3, BC_4$   \\
     &  & $B_x , B_y \in \{50,70,90,110\}$ & $\in \{1,6\}$  \\
      \hline
     \multirow{2}{*}{$u_{tt} - c \nabla^{2}u = 0$} 
     & \multirow{2}{*}{$e=1 ,\, d=0$}  & $c=\{100,150,200,250,300,350\}$ & $BC_1 , BC_2, BC_3$   \\
     &  & $B = [0,0]$ & $\in \{-4,1,6,11\}$  \\
      \hline
     \multirow{2}{*}{$u_{tt} - c \nabla^{2}u + B\nabla u  = 0$} 
     & \multirow{2}{*}{$e=1 ,\, d=0$}  & $c=\{100,200,300\}$ & $BC_1 , BC_2, BC_3 $  \\
     &  & $ B_x , B_y \in \{700,900,1100,1300\}$ & $\in \{1,6\}$  \\
      \hline
     \multirow{2}{*}{$u_{tt} + du_{t} - c \nabla^{2}u = 0$} 
     & \multirow{2}{*}{$e=1 ,\, d=\{200,300\}$}  & $c=\{100,200,300\}$ & $BC_1 , BC_2, BC_3$   \\
     &  & $B = [0,0]$ & $\in \{-4,1,6,11\}$  \\
      \hline
     \multirow{2}{*}{$u_{tt} + du_{t} - c \nabla^{2}u + B\nabla u = 0$} 
     & \multirow{2}{*}{$e=1 ,\, d=\{200,300\}$}  & $c=\{100,150,200,250,300,350\}$ & $ BC_2 , BC_3, BC_4 $  \\
     &  & $ B_x , B_y \in \{900,1100\}$& $\in \{1,6\}$  \\
    \hline
    \end{tabular}}
    \end{center}
    \end{table}
  
\subsection{Automatic Feature Detection}
 
    
    The details of parameters used in Three-dimensional Convolutional Neural Network (3D CNN) and its training are summarized in table S6. 

\begin{table}[htbp]
\begin{center}
\caption{Details of the 3D CNN Model}
\resizebox{\textwidth}{!}{\begin{tabular}{lllll}
\hline
Model              & Parameters             & Settings  & Training Parameters & Settings           \\
\hline
3D CNN  & Num Conv layers      & 2       &   Loss    & Cross-Entropy                \\
  & Conv Kernel      & $3\times3\times3$        &  Classifier    & Softmax                   \\
   & depth Conv layers      & [32,\,64]        & Optimiser    & Adam                  \\ 
& Max Pooling      & $2\times2\times2$        &    Learning Rate  & 0.0017  \\ 
 & Num FC layers      & 3     & &   Scheduled decay    \\ 
 & Width FC layers      & [1728,\,512,\,128,\,*]       &  &        \\ 
\hline
\footnotesize{* Number of classes}
\end{tabular}}
\end{center}
\end{table}

\FloatBarrier

\bibliography{SI.bib}